\documentclass[10pt,journal,compsoc]{IEEEtran}

\ifCLASSOPTIONcompsoc
  \usepackage[nocompress]{cite}
\else
  \usepackage{cite}
\fi
\usepackage{graphicx} 
\ifCLASSINFOpdf
\else
\fi

\usepackage{algorithmic}
\usepackage{amssymb}
\usepackage{booktabs}
\usepackage{multirow}
\usepackage{graphicx}
\usepackage{subfigure}
\usepackage{color}
\usepackage{xcolor}
\ifCLASSOPTIONcompsoc
  \usepackage[caption=false,font=footnotesize,labelfont=sf,textfont=sf]{subfig}
\else
  \usepackage[caption=false,font=footnotesize]{subfig}
\fi
\usepackage{caption}
\begin{document}
%
\title{Flare7K++: Mixing Synthetic and Real Datasets for Nighttime Flare Removal and Beyond}
%
%
%
%

\author{Yuekun~Dai, Chongyi~Li,
        Shangchen~Zhou,
        Ruicheng Feng,\\
        Yihang Luo,
        Chen Change Loy,~\IEEEmembership{Senior Member, IEEE}
\IEEEcompsocitemizethanks{
\IEEEcompsocthanksitem Y. Dai, C. Li, S. Zhou, R. Feng, Y. Luo, and C. C. Loy are with S-Lab, Nanyang Technological  University (NTU), Singapore (E-mail: ykdai005@e.ntu.edu.sg, chongyi.li@ntu.edu.sg, s200094@e.ntu.edu.sg, ruicheng002@e.ntu.edu.sg, c200211@e.ntu.edu.sg, ccloy@ntu.edu.sg)\protect\\
\IEEEcompsocthanksitem C. C. Loy is the corresponding author.}
}

\newcommand{\yuekun}[1]{{\color{black}{#1}}}
\newcommand{\pami}[1]{{\color{black}{#1}}}
\newcommand{\lichongyi}[1]{\textbf{\color{cyan}(CL: {#1})}}
\newcommand{\shangchen}[1]{\textbf{\color{green}(SC: {#1})}}
\newcommand{\ruicheng}[1]{\textbf{\color{orange}(RC: {#1})}}
\newcommand{\cavan}[1]{\textbf{\color{magenta}(CV: {#1})}}



\IEEEtitleabstractindextext{%
\begin{abstract}

\pami{
Artificial lights commonly leave strong lens flare artifacts on the images captured at night, degrading both the visual quality and performance of vision algorithms. Existing flare removal approaches mainly focus on removing daytime flares and fail in nighttime cases.
Nighttime flare removal is challenging due to the unique luminance and spectrum of artificial lights, as well as the diverse patterns and image degradation of the flares.
The scarcity of the nighttime flare removal dataset constraints the research on this crucial task.
In this paper, we introduce Flare7K++, the first comprehensive nighttime flare removal dataset, consisting of 962 real-captured flare images (Flare-R) and 7,000 synthetic flares (Flare7K). 
Compared to Flare7K, Flare7K++ is particularly effective in eliminating complicated degradation around the light source, which is intractable by using synthetic flares alone.
Besides, the previous flare removal pipeline relies on the manual threshold and blur kernel settings to extract light sources, which may fail when the light sources are tiny or not overexposed. 
To address this issue, we additionally provide the annotations of light sources in Flare7K++ and propose a new end-to-end pipeline to preserve the light source while removing lens flares. 
%
Our dataset and pipeline offer a valuable foundation and benchmark for future investigations into nighttime flare removal studies. 
Extensive experiments demonstrate that Flare7K++ supplements the diversity of existing flare datasets and pushes the frontier of nighttime flare removal towards real-world scenarios. 
}

\end{abstract}

}

\maketitle

\IEEEdisplaynontitleabstractindextext

%
\IEEEpeerreviewmaketitle

\ifCLASSOPTIONcompsoc
\IEEEraisesectionheading{\section{Introduction}\label{sec:introduction}}
\else
\section{Introduction}
\label{sec:introduction}
\fi

\begin{figure*}
	\centering

 	\centering
	    \includegraphics[width=1.0\textwidth]{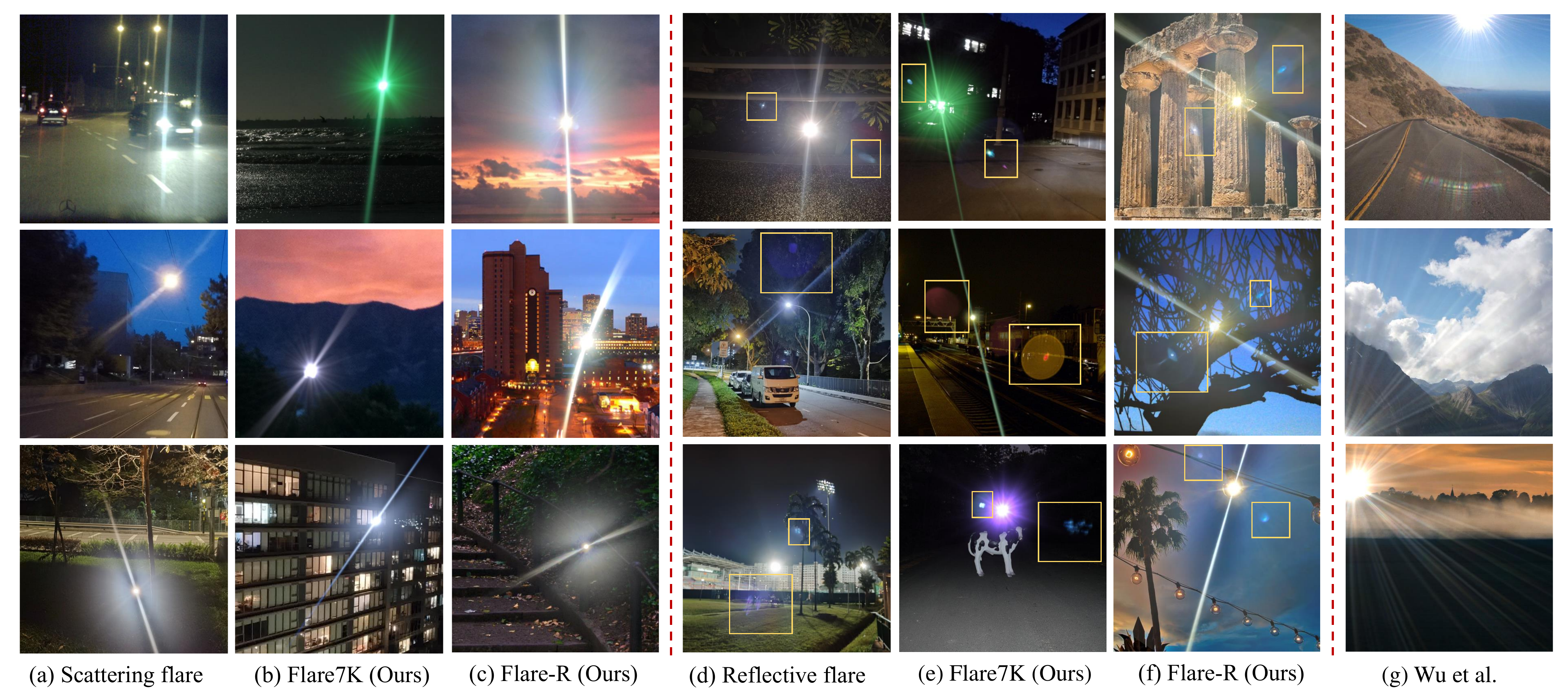}
 	
    %
	\caption{Nighttime photography such as (a) and (d) often suffers from different types of lens flares. In (a), streaks and radial stripes are caused by scattering flare. \pami{In (d), bright blobs and large rings in yellow regions are caused by reflective flare. Columns (b), (c), (e), and (f) depict some of our synthetic nighttime flare-corrupted images with Flare7K++ dataset.} Column (g) shows some flare-corrupted images synthesized by Wu~\textit{et~al.}~\cite{how_to}. Images of reflective flares are taken with the \yuekun{rear} camera of the Huawei P40 smartphone. Parts of images of scattering flares are obtained from Dark Zurich~\cite{darkzurich} and NightOwls~\cite{nightowls} nighttime driving dataset. In contrast to Wu~\textit{et~al.}~\cite{how_to}, our flare data is more similar to real-world nighttime data.} 
	\label{fig:flare_sample}	
\end{figure*}

%
%
%
%
\IEEEPARstart{L}ens flare is an optical phenomenon in which intense light is scattered and/or reflected in an optical system.
It leaves a radial-shaped bright area and light spots on the captured photo. 
The effects of flares are more severe in the nighttime environment due to the existence of multiple artificial lights. 
This phenomenon may lead to low contrast and suppressed details around the light sources, degrading the image's visual quality and the performance of vision algorithms. 
Taking nighttime autonomous driving with stereo cameras as an example, the scattering flare may be misestimated as close obstacles by stereo matching algorithms. 
For aerial object tracking, the bright spots introduced by the lens flare may mislead the algorithm to track flares rather than flying objects~\cite{aerial_tracking}. 

To avoid these potential risks raised by lens flare, the mainstream approaches are to optimize the hardware designs, such as using a specially-designed lens group or applying anti-reflective coatings. 
Although professional lenses can mitigate the flare effect, they cannot solve the inherent problem of flare. 
Besides, fingerprints, dust, and wear in front of the lens often bring unexpected flare that cannot be alleviated by hardware, especially in smartphone and monitor imaging. 
A flare removal algorithm is thus highly desired.

Typical flares can be broadly categorized into three major types: \textit{scattering flare}, \textit{reflective flare}, and \textit{lens orb (a.k.a. backscatter)} \yuekun{\cite{how_to,li2021let,flare_simulation1}}. 
We distinguish these three flares according to their response to the light source movement. 
\textit{Scattering flares} are caused by dust and scratches on the lens. This type of flare produces radial line patterns. 
When moving the lens or the light source, the scattering will always wrap around the light source and keep the same pattern in the captured photo. 
\textit{Reflective flares} are caused by multiple reflections between air-glass interfaces in a lens system \cite{flare_simulation1}. 
Their patterns are determined by the shape of the aperture and lens structure. Such patterns often manifest as a series of circles and polygons on the captured photo \yuekun{\cite{flare_simulation2}}. 
Different from scattering flares, when moving the camera, reflective flares move in the direction opposite to the light source. 
\textit{Lens orbs} are induced by unfocused particles of dust or drops on the lens surface~\yuekun{\cite{gu2009removing}}. 
They are aperture-shaped polygons fixed in the same position of the photograph. 
Only the lens orbs around the light source are lightened, and they do not move with the light source or camera motion. 
Since the position of lens orbs is relatively fixed, this effect is much easier to be removed in a video \yuekun{\cite{li2021let}}. 
Thus, we mainly focus on the removal of scattering flares and reflective flares at nighttime in this study.

Removing nighttime flares is extremely challenging.
First, the flare patterns are diverse, caused by the varied location and spectrum of the light source, unstable defects in the lens manufacture, and random scratches and greasy dirt during daily utilization. 
Second, the dispersion of light at different wavelengths and interference between small optical structures can also lead to rainbow-like halo and colored moiré. 
Although there are some traditional flare removal methods~\cite{automated_removal,auto_removal,auto_removal2}, they mainly concentrate on detecting and wiping off small bright blobs in the reflective flares.
\pami{Recently, some learning-based flare removal methods~\cite{how_to,light_source,rank_1,feng2021removing,feng2023generating} have been proposed for daytime flare removal or removing the flare with a specific type of pattern.
To obtain paired flare-corrupted/flare-free images, Wu~\textit{et~al.}~\cite{how_to} synthesize physically-based flares and capture flare photos with the same cleaned lens in the darkroom. These flare patterns will be overlaid on flare-free images to produce paired data.
Sun~\textit{et~al.}~\cite{rank_1} assume that all flares are generated with the same 2-point star Point Spread Function (PSF). 
Feng~\textit{et~al.}~\cite{feng2021removing,feng2023generating}'s methods mainly focus on removing flare artifacts for under-display cameras.
All these flares generating pipelines assume that lens contaminants or light sources are of a specific type, and lead to relatively homogeneous flare patterns. 
However, for monitor lenses, smartphone cameras, UAVs, and autonomous driving cameras, the fingerprint, daily wear, and dust may function as a grating, thus resulting in streaks (strip-shaped flares) and flares in extremely diverse types of patterns that are obvious at night.
Furthermore, the spectrum of artificial lights often differs significantly from that of the sun, leading to different diffraction patterns that further complicate flare modeling.
The differences between existing synthetic flares and real-world nighttime flares make it difficult to develop models that can generalize well across a variety of nighttime situations.}

\pami{
To facilitate the research on nighttime flare removal, we build a large-scale layered flare dataset \yuekun{with} elaborately designed night flares, called Flare7K++, the first dataset of its kind.
It is composed of 962 real-captured flare images (Flare-R) and 7,000 synthetic flares (Flare7K).
The Flare-R dataset is captured by the rear cameras of Huawei P40, iPhone 13Pro, and ZTE Axon 20 5G.
For each rear lens of the smartphone, we take around 100 photos of flare patterns in the darkroom.
To simulate the daily usage, we wipe the lens with different materials such as finger, silk scarf, and nylon cloth after taking each flare pattern.
Different from the Flare-R, Flare7K is composed of synthetic flare patterns.}
It offers 5,000 scattering and 2,000 reflective flare images, consisting of 25 types of scattering flares and 10 types of reflective flares.
These flare patterns are synthesized based on the observation and statistics of real-world night flares. 
Since scattering and reflective flare are independent, we generate the respective flare data separately. 
Thus, different reflective flares can be added to any scattering flare to obtain pattern diversity.
The 7,000 flare patterns can be randomly added to flare-free images, forming the flare-corrupted and flare-free image pairs that can be used for training deep models. 
In Fig.~\ref{fig:flare_sample}, we present the comparison between our data and the real-world nighttime images. We also present the synthetic data from a recent flare dataset \cite{how_to}. In comparison, our data is more similar to the real-world nighttime images in terms of the flare patterns.
Besides, each scattering flare image in our dataset can be divided into three parts: light source, streak, and glare. 
The separation of flare components makes our dataset more interpretable and manipulatable than previous flare datasets \cite{how_to}. 

\pami
{
This work builds upon our earlier conference version~\cite{dai2022flare7k}. In comparison with the conference version, we have introduced a significant amount of new materials as follows.
1) To overcome the challenge of removing heavy degradation caused by complex diffraction in the lens system, we create a mixing dataset called Flare7K++ that augments the synthetic Flare7K dataset  (conference version) with a new real-captured Flare-R dataset. Utilizing this mixing dataset to train deep models can significantly improve the effectiveness of removing strong degradation.
2) We provide additional light source annotations for all images in Flare7K++ and propose a network to extract light sources from the real-captured flare images.
3) Previous methods such as \cite{dai2022flare7k,how_to} rely on a manual setting of the threshold and blur kernel to extract the light source. It tends to erase the light sources in the final output when light sources are too tiny or not overexposed. Based on our light source annotations, we design a new end-to-end framework for flare removal, which preserves the light source while removing lens flares.
4) We manually label masks for different flare components of test flare-corrupted images and introduce two new metrics, G-PSNR and S-PSNR, to reflect restoration results on glare and streak corrupted regions.
5) To demonstrate the importance of flare removal, we evaluate our method on downstream tasks including stereo matching, optical flow estimation, and semantic segmentation, which are some common tasks in nighttime autonomous driving.
}

\begin{figure}[t]
  \centering
   \includegraphics[width=1.0\linewidth]{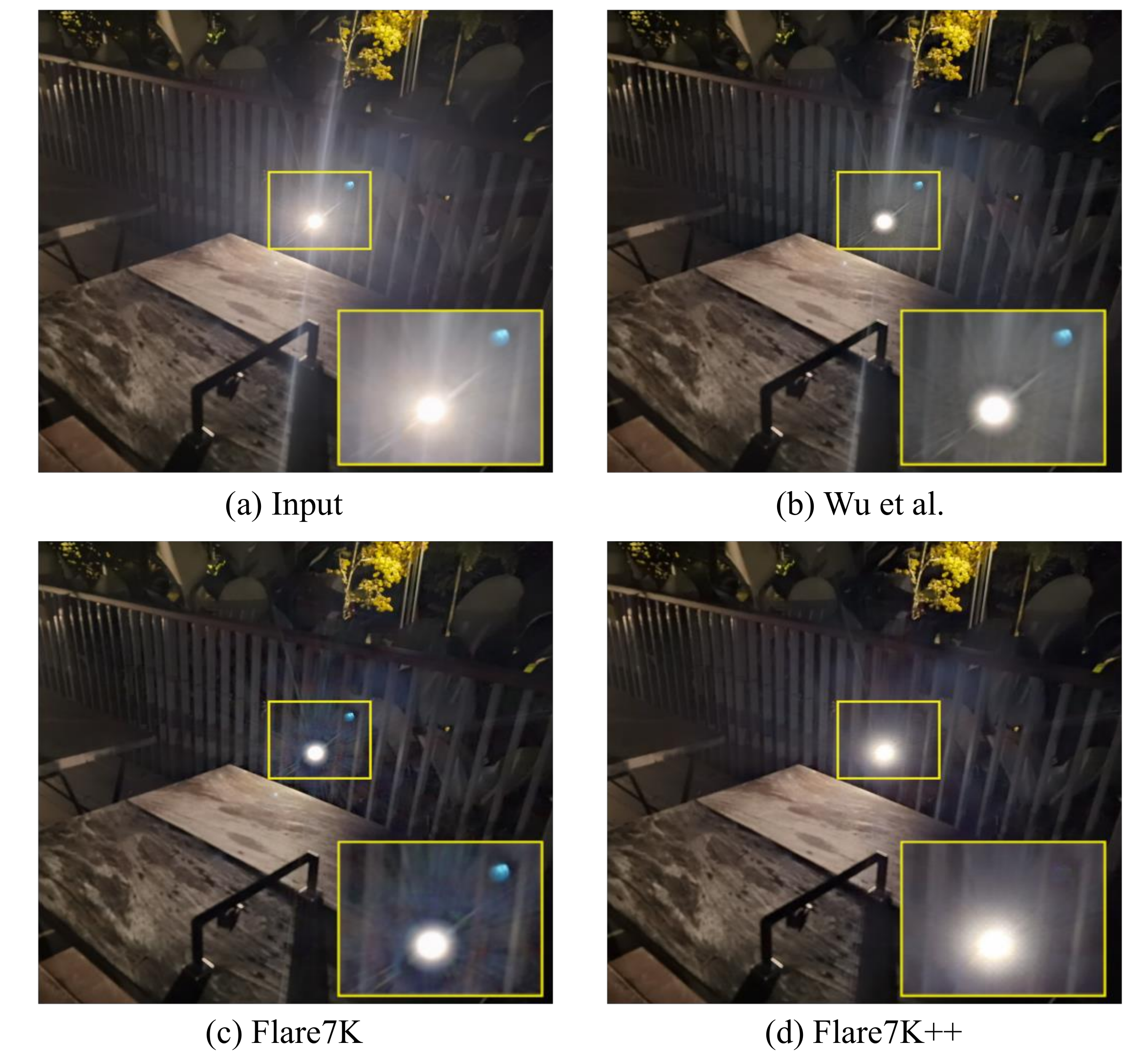}
   \vspace{-7mm}
   \caption{\pami{When dealing with real-world flare-corrupted images, Flare7K outperforms the Wu~\textit{et~al.}~\cite{how_to}'s in terms of eliminating streaks. However, networks trained on the Flare7K dataset tend to produce strong artifacts around the light source. To address this issue, we have developed a new dataset called Flare7K++. Our proposed method, which utilizes the Flare7K++ dataset, effectively removes these artifacts and produces more realistic light sources.} }
   \vspace{-3mm}
   \label{fig:related_work}
\end{figure}

\begin{figure*}[h]
\centering

\includegraphics[width=0.95\textwidth]{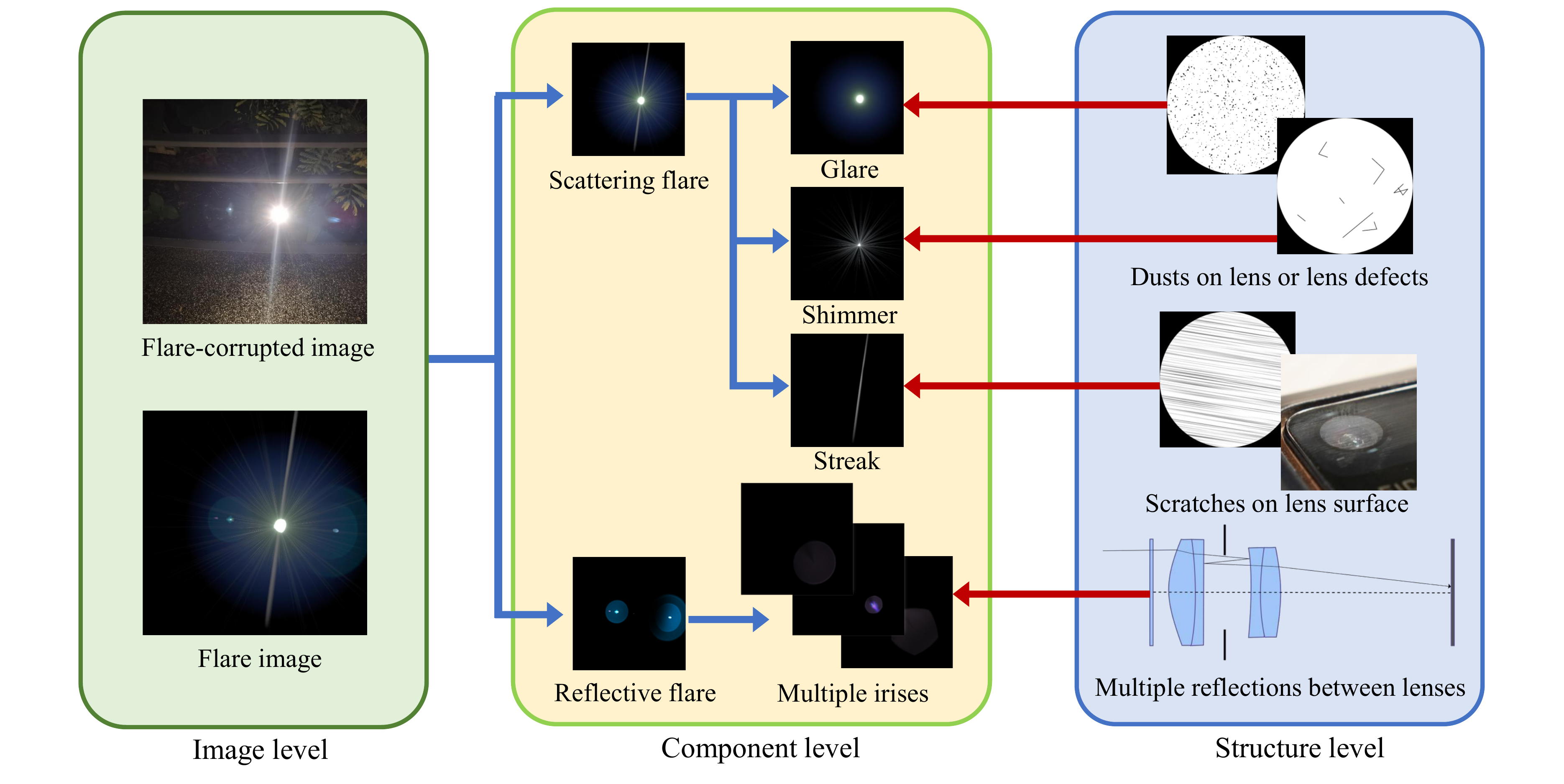}
\caption{Formulation of nighttime lens flare. Lens flare can be viewed as a combination of scattering flare and reflective flare. Multiple reflections between lens surfaces lead to a line of irises pattern, and reflective flare can be viewed as \yuekun{the addition of these irises}.
Scattering flare can be divided into glare, shimmer, and streak. Streak is caused by grating-like scratches in the front of the lenses. Glare and shimmer are brought by the combined action of lens defect and dust on the pupil. Different types of dust make glare and shimmer have different patterns.}
\label{fig:flare_formation} 
\end{figure*}

\section{Related work}
\label{related_work}

\noindent
\textbf{Lens Flare Dataset.}  
Collecting a large-scale captured paired flare-corrupted and flare-free image dataset requires tedious human labor.
To solve this issue, Wu~\textit{et~al.}~\cite{how_to} proposed a semi-synthetic flare dataset, which is the first flare dataset of this kind. It comes with 2,001 captured flare images and 3,000 simulated flare images. 
These flare images can be added to flare-free images to simulate flare-corrupted situations.
However, all the captured flare photos are taken by the same camera and under the same light source within the same distance. 
The homogeneous setting makes the captured flare images look pretty similar and have a limited effect on removing the flares with diverse lenses and light sources.
Besides, Wu~\textit{et~al.}'s synthetic flares also have a considerable gap with real-world nighttime flares, as the comparison shown in Fig.~\ref{fig:flare_sample}. 
Lens flare simulation algorithms~\cite{flare_simulation1, flare_simulation2} have been studied for a long time and are already widely used in visual effects (VFX) of films, animation, and television programs.
With the help of Optical Flares (a plug-in for rendering lens flares in Adobe After Effects), we build a synthetic lens flares dataset Flare7K based on real-world reference images, aiming to solve the problems of domain gap and the lack of diversity.
\pami{Since the complicated artifacts caused by diffraction of the lens contaminants are hard to simulate, a network that is trained only with Flare7K may lead to artifacts around the light source as shown in Fig.~\ref{fig:related_work}.
To address this problem, we additionally collect a real-captured flare dataset Flare-R with contaminated lenses. The combination of these two datasets is named Flare7K++.
}

\noindent
\textbf{Image Flare Removal.}
Prior works mainly focus on veiling glare removal~\cite{HDR_1,HDR_2} of HDR image in the backlit scene and reflective flare removal that involves saturated blobs~\cite{automated_removal,auto_removal,auto_removal2,dai2023nighttime}. 
\yuekun{Due to the lack of paired data that contains scattering flares and diverse reflective flares, deep learning-based methods are restricted.}
Qiao~\textit{et~al.}~\cite{light_source} collected natural flare-corrupted and flare-free images to obtain unpaired flare data. 
Following the idea of Cycle-GAN~\cite{CycleGAN}, Qiao~\textit{et~al.} trained a framework with a light source detection module, a flare generation module, a flare detection module, and a flare removal module. 
Wu~\textit{et~al.}~\cite{how_to} proposed a semi-synthetic flare dataset to synthesize flare-corrupted images. 
With the flare-corrupted image and flare-free image pairs, a pix2pix model based on U-Net~\cite{unet} was trained to restore the flare-free image. 
\pami{
However, Wu~\textit{et~al.}\cite{how_to}'s method relies on a traditional light source extraction algorithm based on threshold and mask feathering, which may fail to work when the light source is not saturated or too small. In this paper, we propose a new end-to-end framework for flare removal that utilizes our provided light source annotations. Our proposed method is capable of effectively removing lens flare while accurately preserving the light source.}

\yuekun{
\noindent
\textbf{Nighttime Defogging and Light Enhancement.}
Multiple scattering of light in fog brings the glare effect around the light source. 
Although the physical principle of this glare effect is different from lens flare, they have similar appearances. 
These properties make nighttime defogging and flare removal share some common ground. 
In the early work of glare effect removal~\cite{shedding_2003}, it is a mainstream method to calculate PSF for each light source and then use deconvolution to recover the light source image. With the wide application of dark channel prior~\cite{dehaze_he}, many nighttime haze removal methods based on statistical priors achieve good performance. 
Li~\textit{et~al.}~\cite{li_nighttime_2015} proposed a new nighttime haze formation model and separated the glare effect by using gray world assumption and smooth glare prior. 
To implement computational efficiency, Zhang~\textit{et~al.}~\cite{nighttime_fast} proposed a fast nighttime haze removal structure using maximum reflectance prior. 
Yan~\textit{et~al.}~\cite{nighttime_yan} noticed that grayscale images are less affected by multiple colors of atmospheric light, thus using the grayscale component to guide a neural network to remove haze.
Zhang~\textit{et~al.}~\cite{nighttime_zhang} designed a new nighttime haze synthesis method that can mimic light rays and object reflectance and then used the synthetic data to train a haze removal network.
Besides nighttime defogging, recent nighttime visibility enhancement methods also focus on light effect suppression. 
In Sharma~\textit{et~al.}'s high dynamic range nighttime image enhancement module~\cite{nighttime_sharma}, a noise and light effect suppression network was introduced to extract low-frequency light effect based on the gray world assumption for the glare-free component. 
For nighttime image enhancement, Zhou~\textit{et~al.}~\cite{zhou2022lednet} presented a learnable non-linear layer to enhance dim regions while avoiding overexposure of light source regions.
All these methods assume that the glare effect is smooth, and hence cannot remove the high-frequency component in scattering flares.
Therefore existing nighttime defogging and visibility enhancement approaches cannot suppress the glare effect effectively.
}

\section{Physics on Nighttime Lens Flare}
\label{physics}
Typical nighttime lens flares usually consist of two types of flares, i.e., \textit{scattering flare} and \textit{reflective flare}, which are complex as they comprise many components including halos, streaks, irises, ghosts, bright lines, saturated blobs, haze, glare, shimmer, sparkles, glint, spike balls, rings, hoops, and caustic. 
In visual effects (VFX), computational photography, optics, and photography, the identical type of components may have different names.
To avoid confusion, we group these names into several common types based on their patterns. 
For instance, sparkles, glints, and spike balls are all radial line-shaped patterns. In this paper, we use shimmer to represent all these types of radial line-shaped components.
To facilitate a better understanding of our proposed dataset, we explain the formation principle of each type of nighttime lens flare below.

\begin{figure*}
	\centering
    \includegraphics[width=1.0\textwidth]{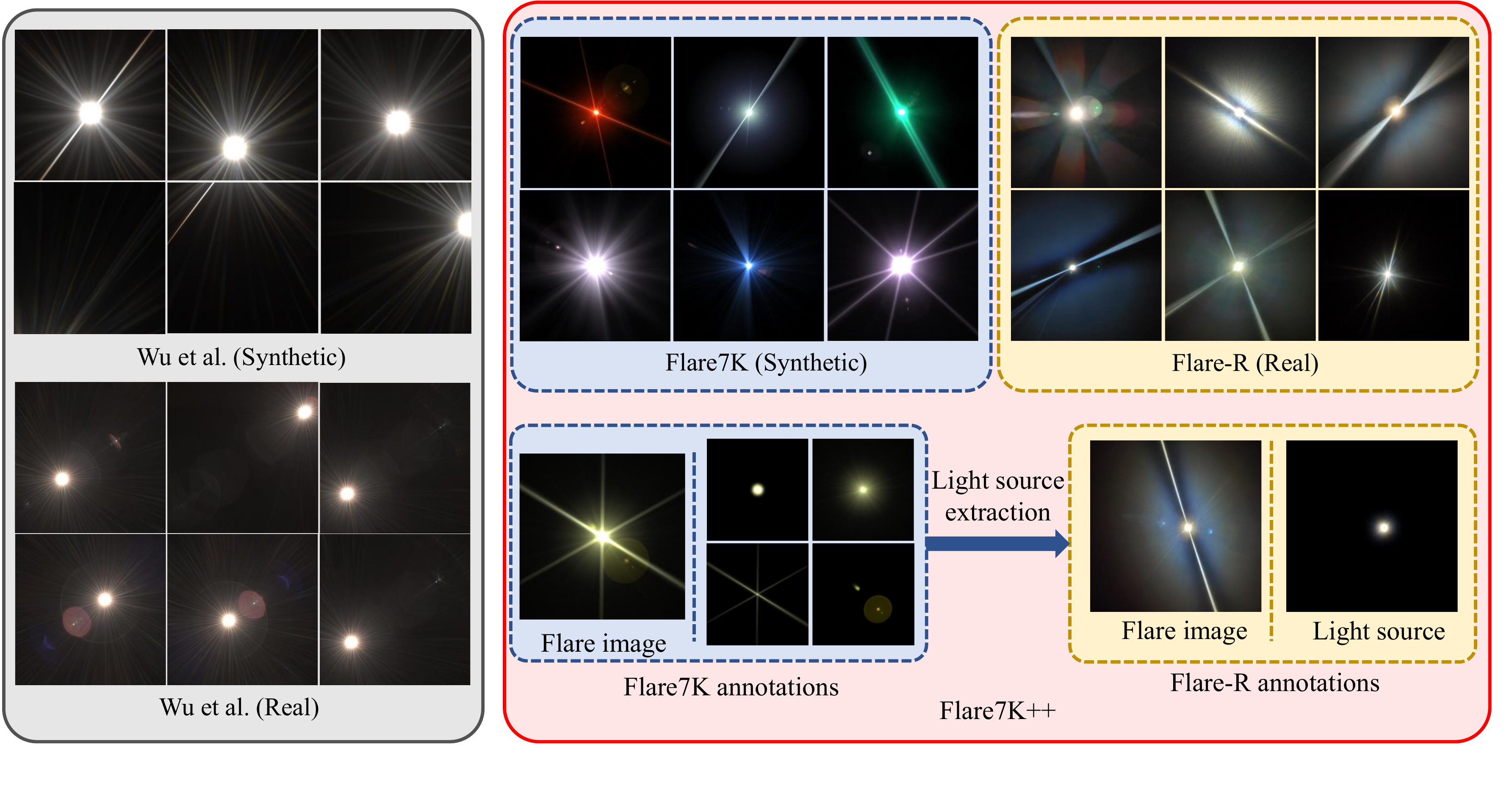}
	\caption{\pami{Comparison between our Flare7K++ dataset and Wu~\textit{et~al.}'s dataset \cite{how_to}. Different from the previous flare dataset, our dataset contains more diverse flare patterns and provides more annotations including light source, glare with shimmer, streak, and reflective flare. The yellow area represents the additional content of this paper compared to the conference version. Its annotations are obtained from a light source extraction module trained on Flare7K dataset.}} 
	\label{fig:compare} 
	
\end{figure*}

\subsection{Scattering Flare} 
The common components in scattering flares can be divided into \textit{glare}, \textit{shimmer}, and \textit{streak}, as shown in Fig.~\ref{fig:flare_formation}. 
\textit{Glare} is a smooth haze-like effect around the light sources, also known as the glow effect~\yuekun{\cite{holladay1926fundamentals}}. 
Even in an ideal lens system, the pupil with a limited radius will still function as a low-pass filter, resulting in a blurry light source. 
Moreover, abrasion or dotted impurities in the lens will lead to the lens' uneven thickness, noticeably increasing the area of the glare effect. 
Besides, dispersion makes the hue of the glare not globally constant. As shown in Fig.~\ref{fig:flare_formation}, the pixels of the glare far away from the light source are bluer than the pixels around the light source.  
During daytime with sufficient illumination, the scene around the light is bright enough to cover the glare effect. 
However, in low-light conditions, the glare is significantly brighter than the scene, hence cannot be ignored for nighttime flare removal.

\textit{Shimmer} (a.k.a., sparkles, glint, spike balls) is a pattern with multiple radial stripes caused by the aperture's shape and line-shaped impurities and lens defects \cite{flare_simulation1}. 
Due to the structure of the aperture, the pupil is not a perfect round, thus producing a star-shaped flare. 
Taking the dodecagon-shaped aperture as an example, diffraction around the edge of the aperture projects a point light source to a dodecagram on the photo. 
\yuekun{Different from the aperture, line-shaped lens defects always lead to uneven shimmer. For the lens flare in the daytime, as a light source with high intensity, the sun will leave bright shimmers over the whole screen.}
In contrast, the intensity of the artificial light is lower and the area of the shimmer is always similar to the glare effect. 
As \yuekun{shimmer} only differs from glare in terms of pattern, it can also be viewed as a high-frequency component of the glare.

\textit{Streaks} (a.k.a., bright lines, stripes) are line-like flares that are significantly longer and brighter than shimmer \yuekun{\cite{rouf2011glare,rank_1}}. 
They often appear in smartphone photography and nighttime driving video. 
Oriented oil stains or abrasions on the front lens may act as grating and cause beam-like PSF. 
During the daytime, streaks are just like brighter shimmer. 
However, in a low-light conditions, even a light source with low intensity may generate streaks across the whole screen. 
Since one cannot always keep a smartphone's lens or vehicle-mounted camera clean, this phenomenon is conspicuous at nighttime.

\vspace{-2mm}
\subsection{Reflective Flare} \label{reflective flare}
\vspace{-1mm}
Reflective flares (a.k.a., ghosting) are caused by reflection in multiple air-glass lens surfaces \cite{flare_simulation1}. 
For a lens system with $n$ optical surfaces, even if the light is exactly reflected twice, there are still $n(n-1)/2$ kinds of combinations to choose two surfaces from $n$ surfaces \yuekun{\cite{how_to,flare_simulation1}}. 
Generally speaking, the reflective flares form a combination of different patterns like circles, polygons, or rings on the image. 
Due to multiple reflections between lenses, it is challenging to synthesize reflective flares in physics. 
To simulate reflective flares, a more straightforward method is to use 2D approaches~\cite{flare_simulation2}. 
Specifically, for 2D reflective flare rendering, since the hoop and ring effect caused by dispersion is not apparent at night, we can abstract the reflective flare as a line of different irises as shown in Fig.~\ref{fig:flare_formation}.
During the process of reflection, if the light path is blocked by the field diaphragm, this would result in a clipping iris. 
In 2D approaches, this effect can be simulated by setting a clipping threshold for the distance between the optical center and the light source. 
If this distance is longer than the clipping threshold, parts of the irises would be clipped proportionally.

Ideally, each iris can be added to the image independently. Moreover, there will not be interference between different irises.
However, in real-world scenes, the neighboring rays are often correlated and generate a triangle mesh. 
To avoid blocking artifacts, Ernst~\textit{et~al.}~\cite{Ernst2005} proposed a way for caustics rendering and introduced a technique for combining and interpolating these irises. 
In our method, since rendering physically realistic caustics increases the difficulty of simulating reflective flare, we use specific caustics patterns to simulate this effect.

\begin{figure*}[!t]
\centering
\includegraphics[width=0.9\textwidth]{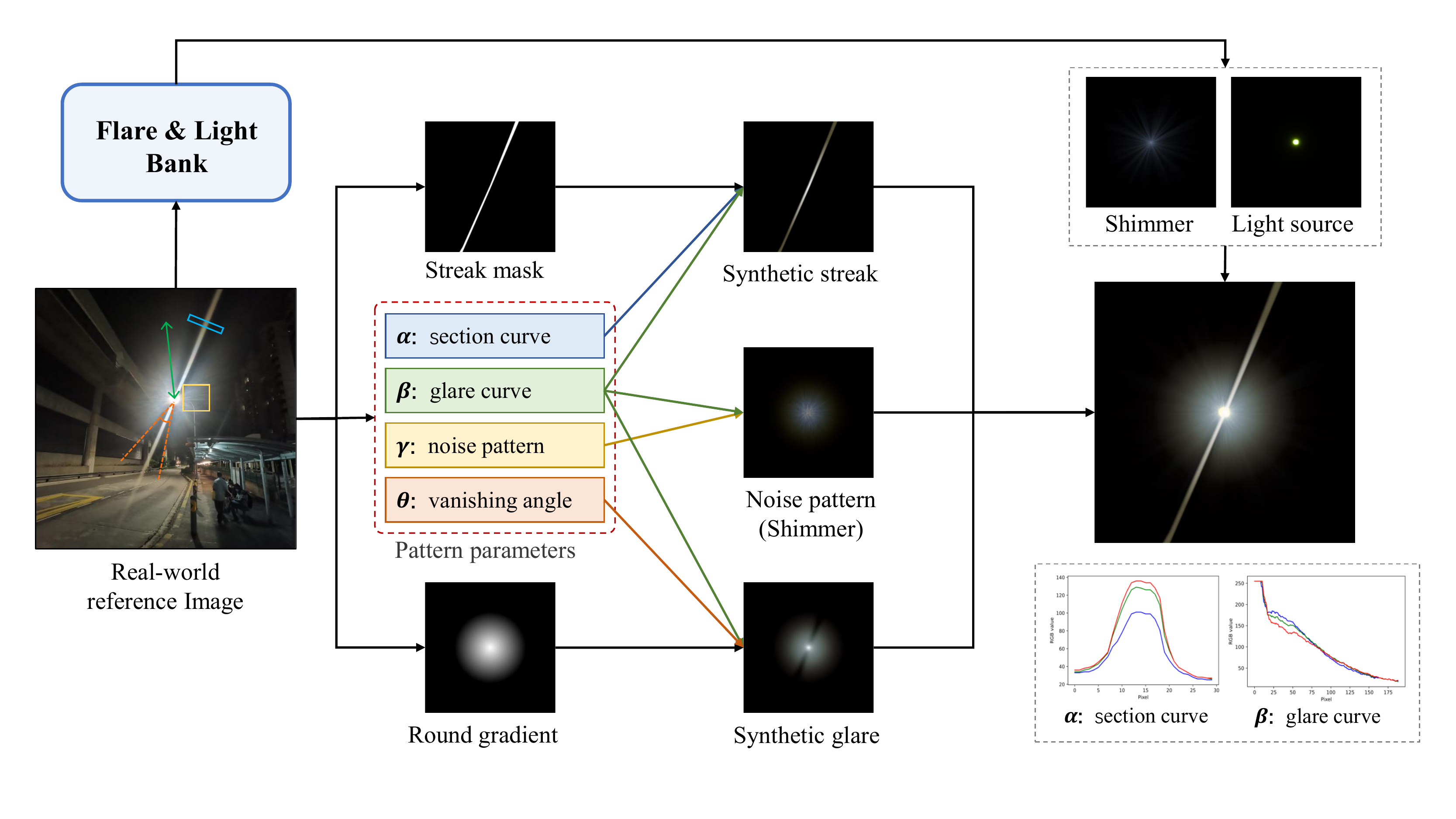}
\caption{The pipeline of scattering flare synthesis in Flare7K dataset. To synthesize scattering flare, we first obtain streak section curve $\alpha$, glare descent curve $\beta$, noise patch near the light source $\gamma$, and the vanishing corner's angle $\theta$ around the streak from the reference image. $\alpha$ and $\theta$ are used to synthesize the glare effect while $\alpha$ and $\beta$ are used to simulate the streak. To simulate the degradation around the light source, we add a blurred fractal noise pattern on the shimmer to create a realistic flare.}
\label{fig:flare_generation} 
\end{figure*}

\section{Flare7K++ dataset}
\label{data_collection}

The only existing flare dataset is the one proposed by Wu~\textit{et~al.} \cite{how_to}, which is mainly designed for daytime flare removal. 
Thus, the streak effect and glare effect that commonly exist in nighttime flares are not considered in Wu~\textit{et~al.}'s dataset. 
In terms of nighttime flares, the patterns are mainly decided by the stains attached on the lens. 
The variety of contamination types makes it difficult for physics-based methods such as Wu~\textit{et~al.}'s approach to collect real nighttime flares by traversing all different pupil functions. 
This results in the lack of diversity and the domain gap between synthetic flares in Wu~\textit{et~al.}'s dataset and that in real-world night scenes. 
\pami{To address the challenges posed by domain gap and lack of diversity, we introduce Flare7K++, a novel dataset that comprises two components: Flare7K, a synthetic flare dataset, and Flare-R, a collection of flare images captured in real-world settings in the dark room.
}
As shown in Fig.~\ref{fig:compare}, our flare patterns are more diverse and closer to real-world nighttime cases.

\subsection{\pami{Flare7K Dataset}} \label{Scattering flare generation}

\begin{table*}[h]
\caption{A comparison between our Flare7K++ dataset \pami{(Flare7K+Flare-R)} and Wu~\textit{et~al.}'s dataset. The `type (s+r)' indicates the number of different patterns of scattering (s) flare and reflective (r) flare. In particular,  we show the numbers as the type of scattering flares + type of reflective flares. \pami{In the Flare-R dataset, the three smartphones we used possess 9 rear cameras which result in 9 types of reflective flares.} Since it is difficult to separate shimmer and glare by definition, we provide glare annotations that also contain the shimmer effect.} 
\label{tab:data_compare_1} 
\centering
\resizebox{0.83\textwidth}{!}{
\begin{tabular}{lcccccccc}

\toprule
\multicolumn{1}{l}{ \multirow{2}*{Dataset} }& \multicolumn{4}{c}{Statistics} &\multicolumn{4}{c}{Annotations}\\
\cmidrule(r){2-5}
\cmidrule(r){6-9}
\multicolumn{1}{l}{}&  number & synthetic & real & type (s+r) & light source & reflective flare & streak& glare\\
\toprule
Wu~\textit{et~al.}~\cite{how_to}             &  5,001& 3,000& 2,001& 2+1    &   $\times$&$\times$&$\times$&$\times$\\
Flare7K (ours)                    &  7,000& 7,000& 0& 25+10      &   $\checkmark$&$\checkmark$&$\checkmark$&$\checkmark$\\
Flare-R (ours)                   &  962& 0& 962& 962+9      &   $\checkmark$&$\times$&$\times$&$\times$\\
\hline
\end{tabular}
}
\end{table*}

To synthesize Flare7K, we take hundreds of nighttime flare images with different types of lenses (smartphone and camera) and various light sources as reference images. 
We aggregate the captured images and summarize the scattering flares as 25 typical types based on their patterns.
For each type of flare, we generate 200 images with different parameters such as the glare's radius, the streak's width, etc. 
Since reflective flares are directly related to the type of lens group, we also capture some video clips using different cameras as references for reflective flare synthesis.
By referring to these real-world nighttime flare videos, we design a group of irises for each specific kind of camera and synthesize 10 typical types of reflective flares. 
We finally obtain 5,000 scattering flares and 2,000 reflective flares.
Since flare rendering is relatively mature, we choose to directly use the plug-in Video Copilot's Optical Flares in Adobe After Effects to generate customized flares. 
Fig.~\ref{fig:flare_generation} presents our scattering flare synthesis pipeline. We separate the lens flare into four components including shimmer, streak, glare, and light source. For each component, we analyze the parameters like glare's radius range and color-distance curve in reference images. Then, we use Adobe After Effect to synthesize flare templates.

\noindent
\textbf{Glare Synthesis.} 
From the reference flare-corrupted images of each type, we first plot the relationship between the RGB value of the pixel and its distance to the light source as the glare curve. 
Divided by the glare's radius, such a color-distance relationship can be viewed as a color correction curve. 
Applying this curve to a round gradient pattern with the glare's radius can produce the glare effect of this type of flare. 
Since the region's luminance around the streak sometimes becomes weaker than the normal area, we measure the vanishing angle manually and use a feathered mask around the streak to decrease the opacity of glare in these areas. 
The angle of this missing corner is set to a variable to cover more cases while generating this type of scattering flare. 

\noindent
\textbf{Streak Synthesis.} \label{Streak Synthesis}
In Sun~\textit{et~al.}~\cite{rank_1}'s work, it assumes that the streak is always generated with a 2-point star PSF. 
However, the streak effect is not even symmetric, and one side is often much sharper than the other side.
To imitate this effect, we manually draw a mask for each type of streak in Adobe After Effect and set the width as a variable.
Then, we plot the RGB value of the streak's section and glare section and use this curve to colorize the streak and blur the mask's edge.
The blur size for each edge is derived from the section curve's half-life value.

\noindent
\textbf{Shimmer Synthesis.} \label{Shimmer Synthesis}
As for shimmer, we use the shimmer template of Optical Flares and adjust the parameters until it roughly matches the flare of the image. 
In the area around the light source, the image often suffers from strong degradation that is challenging to be simulated by Optical Flares. 
If we suppose our lens flare is smooth, this degradation would be separated as part of the deflared image. 
Thus, we use Adobe After Effect's default plug-in fractal noise to generate a noise patch and then add the radial blur effect to it, thus creating a radial noise pattern. 
This pattern will be added to a shimmer template of Optical Flares to compose a realistic shimmer.

\noindent
\textbf{Light Source Synthesis.} \label{Light source Synthesis}
We apply thresholding on flare-corrupted images to obtain a group of overexposed tiny shapes.
To simulate the light source's glow effect, we apply another plug-in named Real Glow on different tiny shapes in Adobe After Effects. 
To ensure that only the light source region is overexposed, the light source is made larger than the glare's overexposure part. 
Then, it is added to the flare with screen blend mode. 
This mode ensures that the overexposed region is not expanded and brings realistic visual results.
\pami{These synthetic light sources will be used as light source annotations in our training pipeline in Section~\ref{sec:training_pipeline}.}

\begin{figure*}[!t]
\centering
\includegraphics[width=0.9\textwidth]{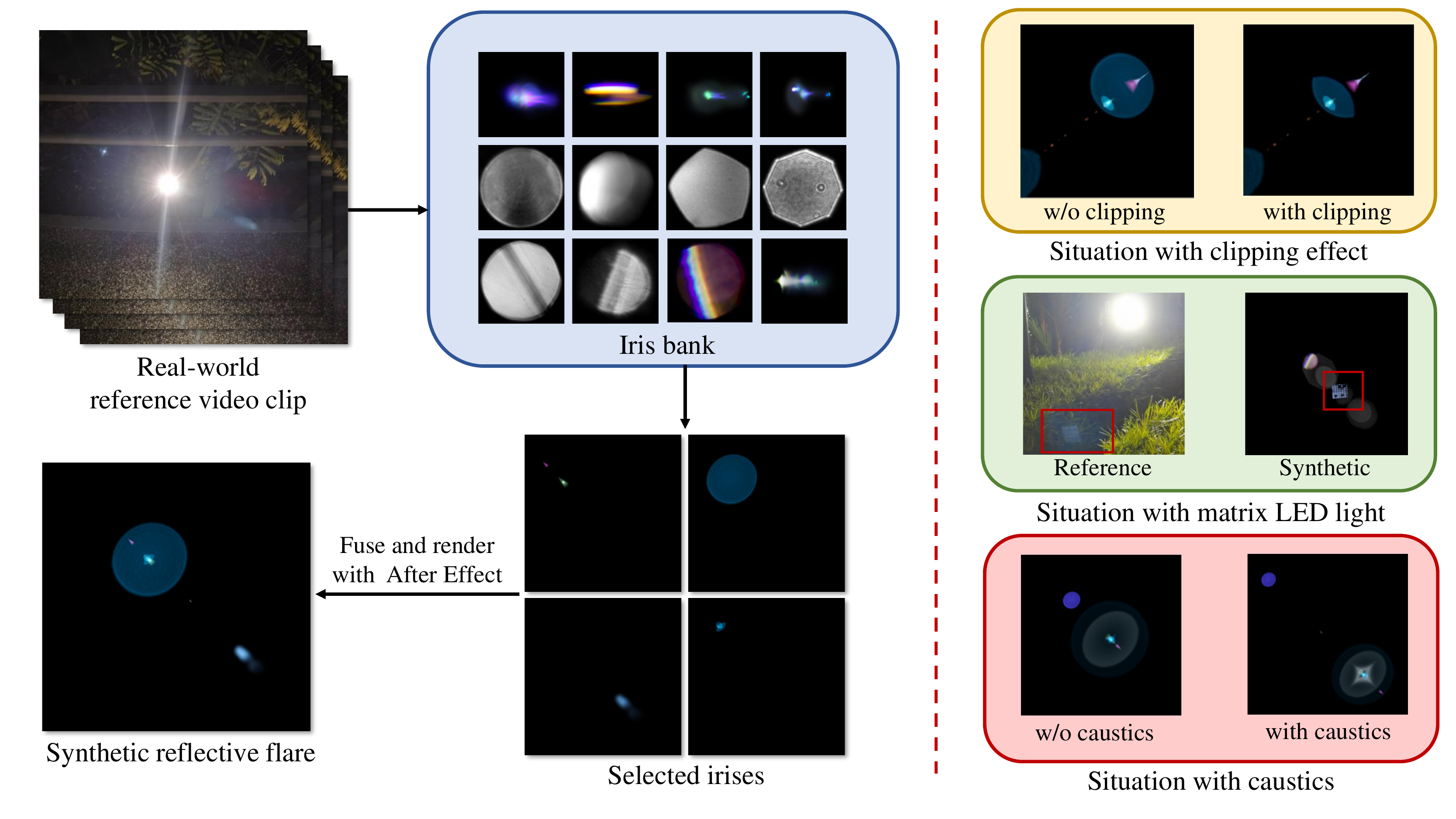}
\vspace{-2mm}
\caption{\yuekun{The pipeline of reflective flare synthesis in Flare7K dataset. Since clipping effects and caustics are not obvious in a single image, we capture video clips as references. While synthesizing reflective flares, we first filter most similar irises from Optical Flares Plug-in's iris bank. Then, we manually adjust the position, size, and color of these irises to fit the reference. Finally, these irises are fused to create a reflective flare template in Adobe After Effect. For some special cases like caustics, matrix light, or clipping effect, the details are presented in Section \ref{Reflective_flare_generation}.} }
\vspace{-4mm}
\label{fig:reflective_flare_generation} 
\end{figure*}

\noindent
\textbf{Reflective Flare Synthesis.} \label{Reflective_flare_generation}
For reflective flares, the plug-in Optical Flares' Pro Flares Bundle contains 51 kinds of different captured high-quality iris images that can serve as the iris bank.
\yuekun{
While comparing with the reference video, we pick the most similar irises and manually adjust their size and color with the Optical Flares plug-in.
Since the distances from different irises to a light source are always proportional, we follow the plug-in's pipeline to set different iris components in a line with proportional distance to a light source. 
After that, we can obtain a reflective flare template.

For some special types of reflective flares, we also consider flares' dynamic triggering mechanisms like caustics and clipping effect. 
These phenomena will happen when the light source's position on the image is far from the lens' optical center.
As stated in Section \ref{reflective flare}, the caustics phenomenon is caused by interference between different irises.
To simulate this effect, we use Optical Flares' default caustics template to generate a caustics pattern in the center of the iris.
To simulate the dynamic triggering effect, the opacity of this caustics pattern is set to be proportional to the distance between the iris and the light source.
As for the clipping effect, it is generated when the reflected light path is blocked by more than two lenses' apertures.
It can be viewed as the intersection of two irises.
Thus, when the iris-light distance is larger than the clipping threshold, we start to erase parts of the iris by using another iris as a mask.
This iris will only serve as a mask and will not be rendered.

In nighttime situations, matrix LED light is common and may bring lattice-shaped reflective flare. 
To imitate this effect, we synthesize some irises in the shape of the lattice as shown in Fig~\ref{fig:reflective_flare_generation}.
Compared to the previous dataset \yuekun{\cite{how_to}}, these designs reflect real-world nighttime situations better.
}

\pami{
\subsection{Flare-R Dataset}

\pami{Since the synthetic dataset does not contain complicated degradation caused by diffraction and dispersion in the lens system, we also capture a real-world flare dataset, Flare-R, with 962 flare patterns.
Different from Wu~\textit{et~al.}'s capturing method, we reproduce common lens contaminants in daily use to enrich our dataset's diversity.
}
We dip different types of liquid including water, oil, ethyl alcohol, and carbonated drinks on the lens surface and wipe it with our fingers and different types of clothes.
After each wipe, we capture a new lens flare image.
We disable the automatic white balance of the phone cameras and obtain different-colored lens flare images by changing the color temperature of the light source.
The Huawei P40, iPhone 13 Pro, and ZTE AXon 20 5G all have three rear lenses with different focal lengths. 
For each lens, we vary the distance from the camera to the light source and capture approximately 100 images.
In total, we collected 962 flare images covering almost all common situations. 
Figure~\ref{fig:compare} presents examples to show the difference between synthetic and real flare patterns.
Unlike the synthetic Flare7K data, obtaining the light source annotations for the real-captured flares is difficult.
To solve this problem, we train a network with Flare7K and its light source annotations to extract the light source from the real flare images.
We use this network to process all images in the Flare-R dataset to automatically get the light source annotations.
For very few failure cases where reflective flares are not removed, affecting the accuracy of light source extraction, we manually erase erroneous bright spots to obtain the light source annotations.
The details of light source extraction are provided in Section~\ref{light_source_extraction}.
}

\begin{figure*}[!t]
\centering
\includegraphics[width=1.0\textwidth]{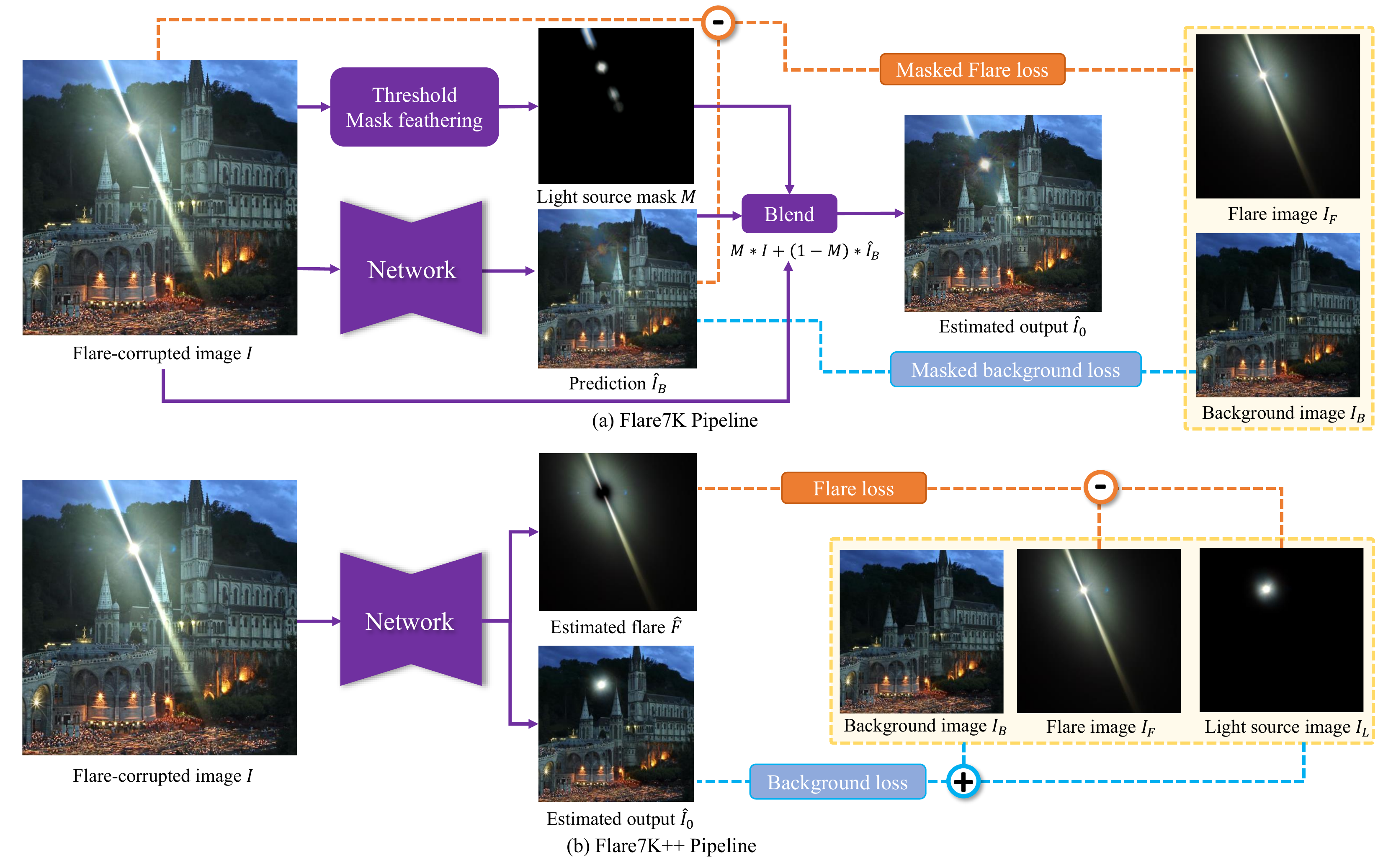}
\caption{
\pami{
The comparison of the previous pipeline (a) applied by  Wu~\textit{et~al.}, Flare7K\cite{how_to,dai2022flare7k} and our current nighttime flare removal training and inference process (b). In the figure, purple lines represent the inference pipeline, while the yellow boxes indicate a sampled training batch.
In the training stage of our proposed method (b), paired flare and light source images will be randomly selected from the Flare7K++ dataset. The flare image will be added to the background image to synthesize the flare-corrupted image. Different from the previous method~\cite{how_to,dai2022flare7k} that outputs a 3-channel flare-free prediction, our network produces a 6-channel output that includes a 3-channel flare-free image and a 3-channel flare image. We use the light source image added to the background image to generate the flare-free ground truth image. We also calculate the difference between the flare image and the light source image to produce the ground truth for the flare image. These ground truth images are used to supervise the network during training. }} 
\label{fig:flare_removal_pipeline} 
\end{figure*}

\begin{figure}[t]
  \centering
   \includegraphics[width=1.0\linewidth]{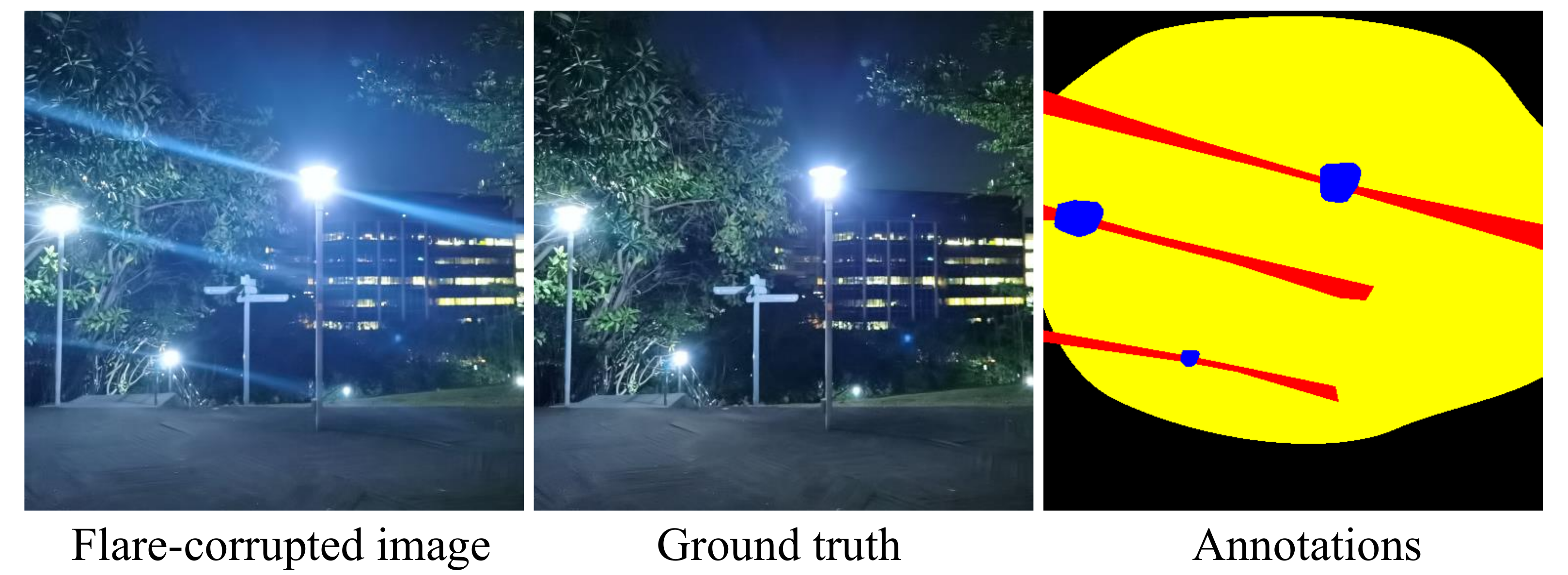}
   \caption{\pami{Typical example sampled from the test dataset. For each test image, we provide a corresponding flare segmentation map. Glare, streak, and light sources are labeled in yellow, red, and blue, respectively. S-PSNR means the PSNR in the red region and G-PNSR represents the PSNR in the sum of the yellow and red regions.} }
   \label{fig:test_type}
\end{figure}

\subsection{Comparison with Existing Flare Dataset}
We compare the differences between our dataset and Wu~\textit{et~al.}'s dataset in Table~\ref{tab:data_compare_1}. 
The comparison shows that our new dataset offers richer patterns and annotations, which benefit broader applications, such as lens flare segmentation and light source extraction.
The details of extended applications are presented in Section~\ref{sec:extended_applications}. 
In addition, Fig.~\ref{fig:compare} shows that our new dataset is more representative of real-world nighttime flares in terms of color and pattern.

\subsection{Paired Test Data Collection}
Since there is no publicly available nighttime flare removal test dataset, we collect new real nighttime flare paired data for full-reference evaluations.
For most well-designed lenses, the flares of a nighttime scene are mainly caused by the stains on the lens's surface (or scratches on the windshield for nighttime driving). 
To reproduce these flares, we use fingers and a cloth to wipe the front lens of the camera to mimic common stains. 
After that, we use lens tissue to clean the front lens slightly to obtain flare-free ground truth.
The action of cleaning may still cause a tiny misalignment of the paired images. 
Thus, we align the paired images manually and obtain \yuekun{100} pairs of real-world flare-corrupted/flare-free images as our real-world test dataset. 
\pami{
Since the ground truth may still be influenced by the slight flares brought by the lens's defects, global PSNR cannot fully reflect the performance of flare removal methods. 
To address this problem, we manually labeled masks for all streak, glare, and light source regions as shown in Fig.~\ref{fig:test_type}.
Masked PSNR can be used to evaluate the restored results in the regions of different components of flares.
We name the masked PSNR of glare and streak regions G-PSNR and S-PSNR.

}

\begin{figure*}[t]
    \centering
    \includegraphics[width=1.0\textwidth]{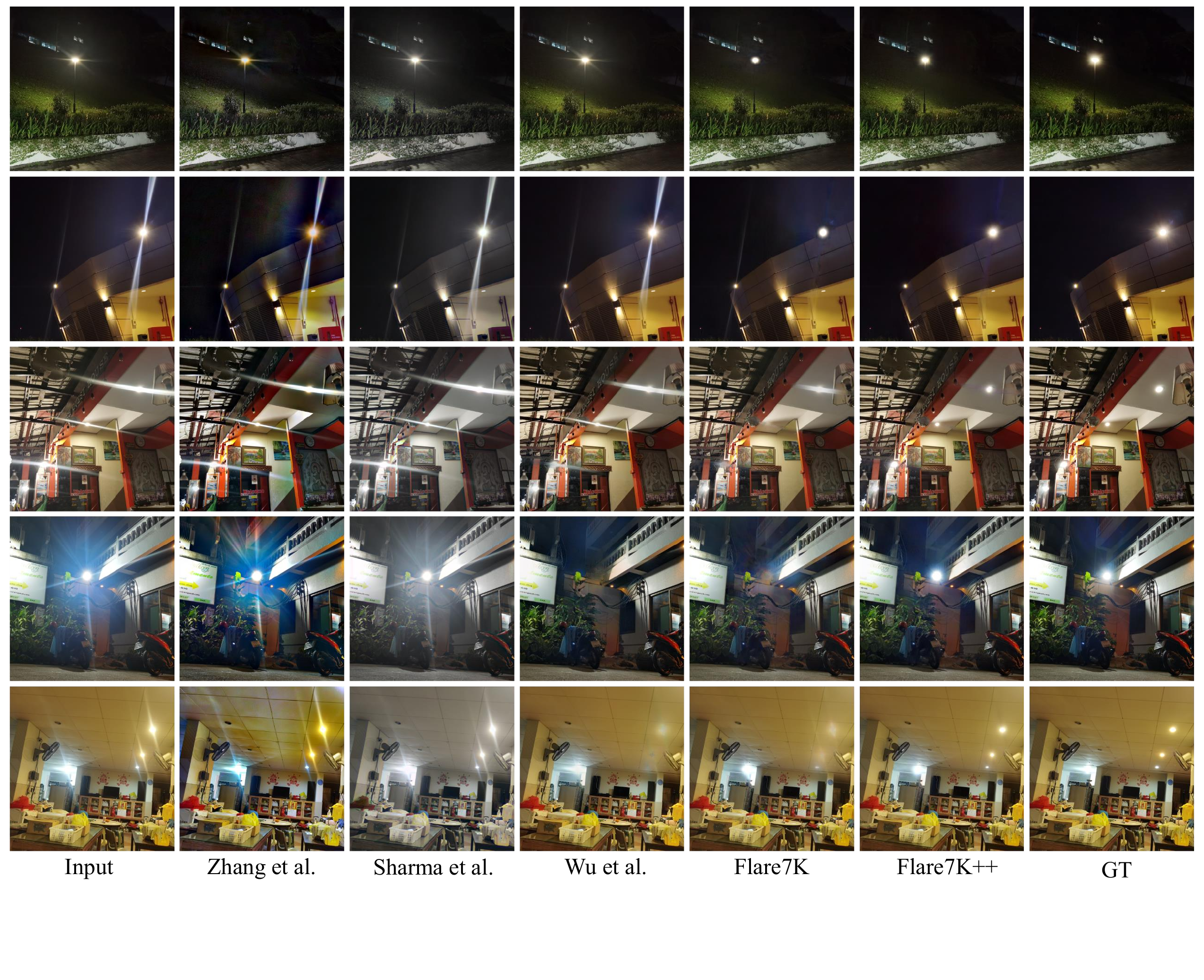}
    \vspace{-20mm}
    \caption{\pami{Visual comparison of flare removal on real-world nighttime flare images. Flare7K represents training network with pipeline proposed by Wu~\textit{et~al.}~\cite{how_to}. As shown in the last two lines, when the light sources are not bright or large enough, the threshold and mask feathering module may fail to reconstruct the light source. Our new dataset and the proposed method can help achieve more realistic light sources and eliminate the streak effect better. }} 
    \label{fig:result_comparison}
\end{figure*}

\pami{
\section{Proposed Method}

In the inference pipeline proposed by Wu~\textit{et~al.}~\cite{how_to}, the network outputs a flare-free prediction that lacks a light source. 
The flare-corrupted image is then processed with a manual threshold value and blur kernel to extract the light source mask, as shown in Fig.~\ref{fig:flare_removal_pipeline}. 
However, the streak region may still get overexposed, leaving parts of the flare on the image. 
Moreover, thresholding-based algorithms may fail to segment light sources that are not bright enough, and blurring operations may inadvertently erase tiny light sources. 
Consequently, it is desired to have a method that can accurately preserve light sources while removing flare artifacts. 
To address these challenges, we introduce an end-to-end training pipeline that uses our light source annotations, ensuring the reliable preservation of light sources during flare removal.

\subsection{Training Pipeline} \label{sec:training_pipeline}
As shown in Fig.~\ref{fig:flare_removal_pipeline}, our previous Flare7K pipeline~\cite{how_to,dai2022flare7k} takes a flare-corrupted image $I$ as input and outputs a flare-free image $\hat{I}_B$. 
Then, the estimated flare image $\hat{I}_F$ will be predicted by calculating the difference between the input $I$ and the output $\hat{I}_B$.
Threshold and erosion operation will be applied on flare-corrupted image $I$ to obtain the mask $M_0$.
A masked loss of the estimated flare and flare-free image is calculated on the region outside the mask to drive the network to focus on reconstructing the details outside the light source.
Thus, this previous method produces a flare-removed image without any light source and pastes the light source back by post-processing.

Our new Flare7K++ pipeline modifies the framework to directly predict a 6-channel output with 3 channels as a flare-free image $\hat{I}_0$ and the other 3 channels as a flare image $\hat F$.
To ensure these two images can be added to get the original input, a reconstruction loss is used to supervise the final output. 
Different from the previous flare dataset~\cite{how_to}, our Flare7K++ dataset provides light source annotations.
As shown in Fig.~\ref{fig:flare_removal_pipeline}, light source image $I_L$ is added to the background images $I_B$ to synthesize the ground truth of flare-free images $I_0$.
Then, the difference between the flare image $I_F$ and light source image $I_L$ is calculated to supervise the estimated flare $\hat F$.
This allows us to train end-to-end networks without manually setting thresholds to extract the light source.
Thus, our pipeline enables a network to distinguish overexposed light sources and overexposed streaks.
Even for non-saturated light sources, our learning-based method can also produce accurate output.

\begin{figure*}[t]
    \centering
    \includegraphics[width=1.0\textwidth]{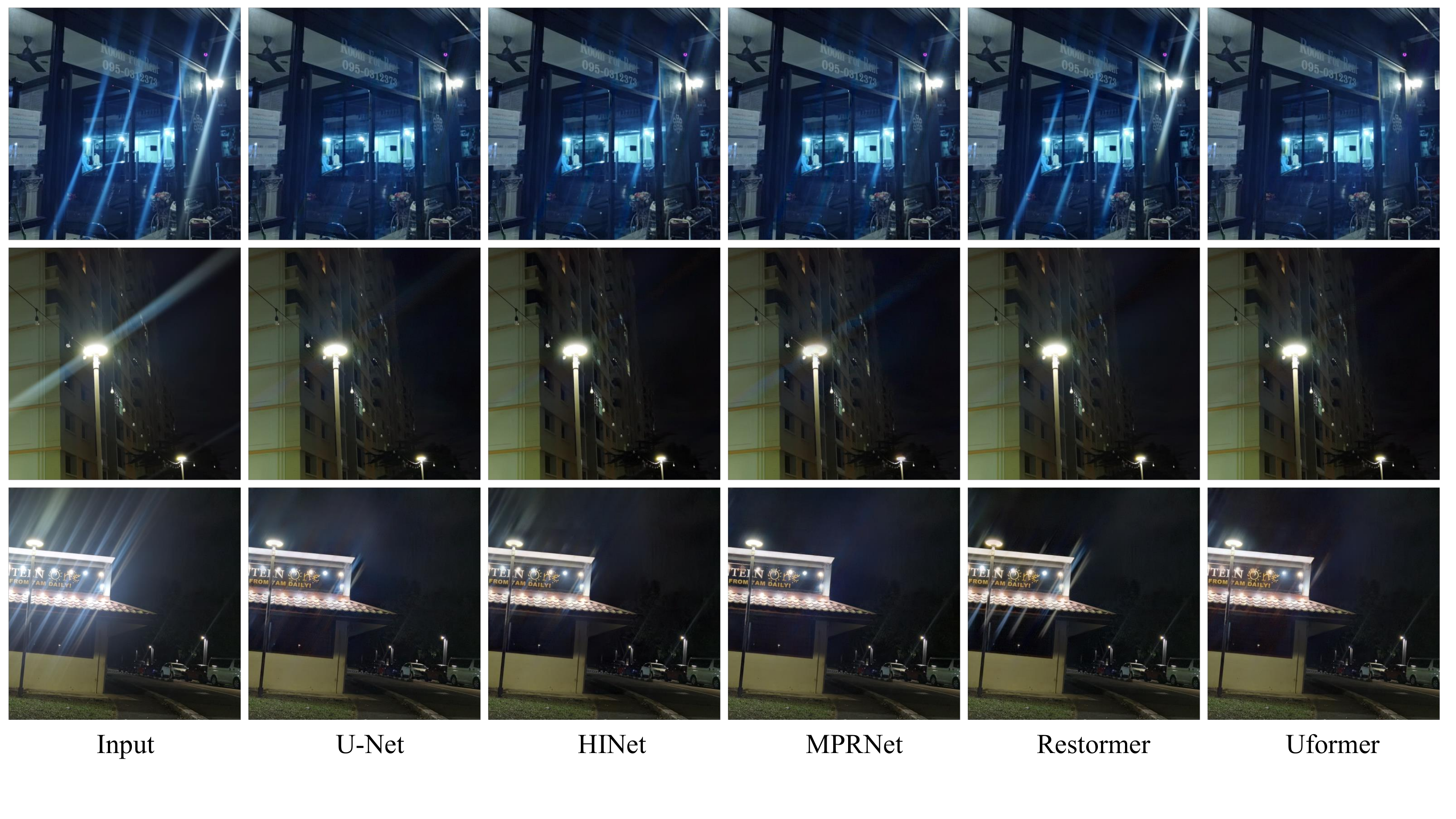}
    \vspace{-14mm}
    \caption{\pami{Visual comparison on our real-world test images for different image restoration networks trained on our Flare7K++ dataset. These networks include U-Net~\cite{unet}, HINet~\cite{HINet}, MPRNet~\cite{MPRNet}, Restormer~\cite{Restormer}, and Uformer~\cite{Uformer}.}} 
    \label{fig:network_comparison}
\end{figure*}

\begin{table*}[th]
\vspace{-1mm}
\caption{\yuekun{
Quantitative comparison of synthetic and real nighttime flare-corrupted data. 
In the experiments, Wu~\textit{et~al.}~\cite{how_to} use U-Net~\cite{unet} as the backbone network and Flare7K's baseline applies Uformer~\cite{Uformer}. Experiments show that our Flare7K++ dataset has better performance on real nighttime flare-corrupted images.
The benchmark of the image restoration methods for nighttime flare removal is listed on the right part of the table. 
"*" denotes models with reduced parameters due to the limited GPU memory. 
It is expected that their original models would perform better.
}}

\label{tab:PSNR_SSIM} 
\centering
\resizebox{1.0\textwidth}{!}{
\begin{tabular}{llccccccccccc}
\toprule
\multirow{2}{*}{Metric\textbackslash{}Method} & \multirow{2}{*}{} & \multirow{2}{*}{Input} & \multicolumn{4}{c}{Previous work}                                                                                                                     & \multicolumn{1}{l}{}                 & \multicolumn{5}{c}{Network trained on Flare7K++}       \\ \cmidrule{4-7} \cmidrule{9-13} 
                                              &                   &                        & Zhang~\cite{nighttime_zhang} & Sharma~\cite{nighttime_sharma} & Wu~\cite{how_to} & \multicolumn{1}{l}{Flare7K baseline} &  & U-Net~\cite{unet} & HINet~\cite{HINet} & MPRNet*~\cite{MPRNet} & Restormer*~\cite{Restormer} & Uformer~\cite{Uformer} \\ \midrule
PSNR$\uparrow$                                &                   & 22.561                 & 21.022                                               & 20.492                                                 & 24.613                                   & 26.978                               &  & 27.189                                  & 27.548                                                             & 27.036                                                                & 27.597                                                                      & \textbf{27.633 }                                                                \\ \midrule
SSIM$\uparrow$                                &                   & 0.857                  & 0.784                                               & 0.826                                                 & 0.871                                   & 0.890                                &  & 0.894                                   & 0.892                                                              & 0.893                                                                 & \textbf{0.897}                                                                       & 0.894                                                                  \\ \midrule
LPIPS$\downarrow$                             &                   & 0.0777                 & 0.1738                                               & 0.1115                                                 & 0.0598                                   & 0.0466                               &  & 0.0452                                  & 0.0464                                                             & 0.0481                                                                & 0.0447                                                                      & \textbf{0.0428}                                                                 \\ \midrule
G-PSNR$\uparrow$                              &                   & 19.556                 & 19.868                                                    & 17.790                                                      &  21.772                                      & 23.507                               &  & 23.527                                  & \textbf{24.081}                                                             & 23.490                                                                 & 23.828                                                                      & 23.949                                                                 \\ \midrule
S-PSNR$\uparrow$                              &                   & 13.105                 &     13.062                                                &   12.648                                                    &      16.728                                   & 21.563                               &  & 22.647                                  & \textbf{22.907}                                                             & 22.267                                                                & 22.452                                                                      & 22.603  \\ \midrule
\end{tabular}
}
\end{table*}

\subsection{Loss Function}
In the training process, paired flare image and light source image are randomly sampled from the Flare7K++ dataset.
The light source image $I_L$ is added to the background image $I_B$ to obtain the ground truth of the flare-free image $I_0$.
Then, the difference between the flare $I_F$ and light source $I_L$ is used as the ground truth of the flare image $F$.
Since our flare images and Flickr background images are all gamma-encoded, we follow Wu~\textit{et~al.}'s pipeline and apply an approximate inverse gamma correction curve with $\gamma$ sampled from [1.8,2.2] to linearize the images before each addition and subtraction operation.
Our flare removal network can be defined as $\Phi$, and it takes the flare-corrupted image $I$ as input. Then, the estimated flare-free image $\hat{I}_0$ and flare image $\hat{F}$ can be expressed as:
\begin{equation}
\label{eq1}
\hat{I}_{0},\hat{F}=\Phi{(I)}.
\end{equation}

Like previous work~\cite{how_to,dai2022flare7k}, flare and background images are supervised using $L_1$ loss and perceptual loss $L_{vgg}$. The background image loss $L_B$ can be written as:
\begin{equation}
\label{eq2}
L_B=L_1(\hat{I}_0,I_0)+L_{vgg}(\hat{I}_0,I_0),
\end{equation}
which shares the same expression with the flare loss $L_F$.

The reconstruction loss $L_{rec}$ is defined as:
\begin{equation}
\label{eq3}
L_{rec}=|I-Clip(\hat{I}_0 \oplus \hat{F})|,
\end{equation}
where $\oplus$ means the addition operation in the linearized gamma-decoded domain with previously sampled $\gamma$. 
Then, the addition is clipped to the range of [0,1]. Overall, the final loss function aims to minimize a weighted sum of all these losses:
\begin{equation}
\label{eq4}
\mathcal{L}=w_1L_B+w_2L_F+w_3L_{rec},
\end{equation}
where $w_1,w_2,w_3$ are respectively set to 0.5, 0.5, and 1.0 in our experiments.

}

\section{Experiments}
\label{experiments}

\begin{figure*}
	\centering
	\includegraphics[width=1.0\linewidth]{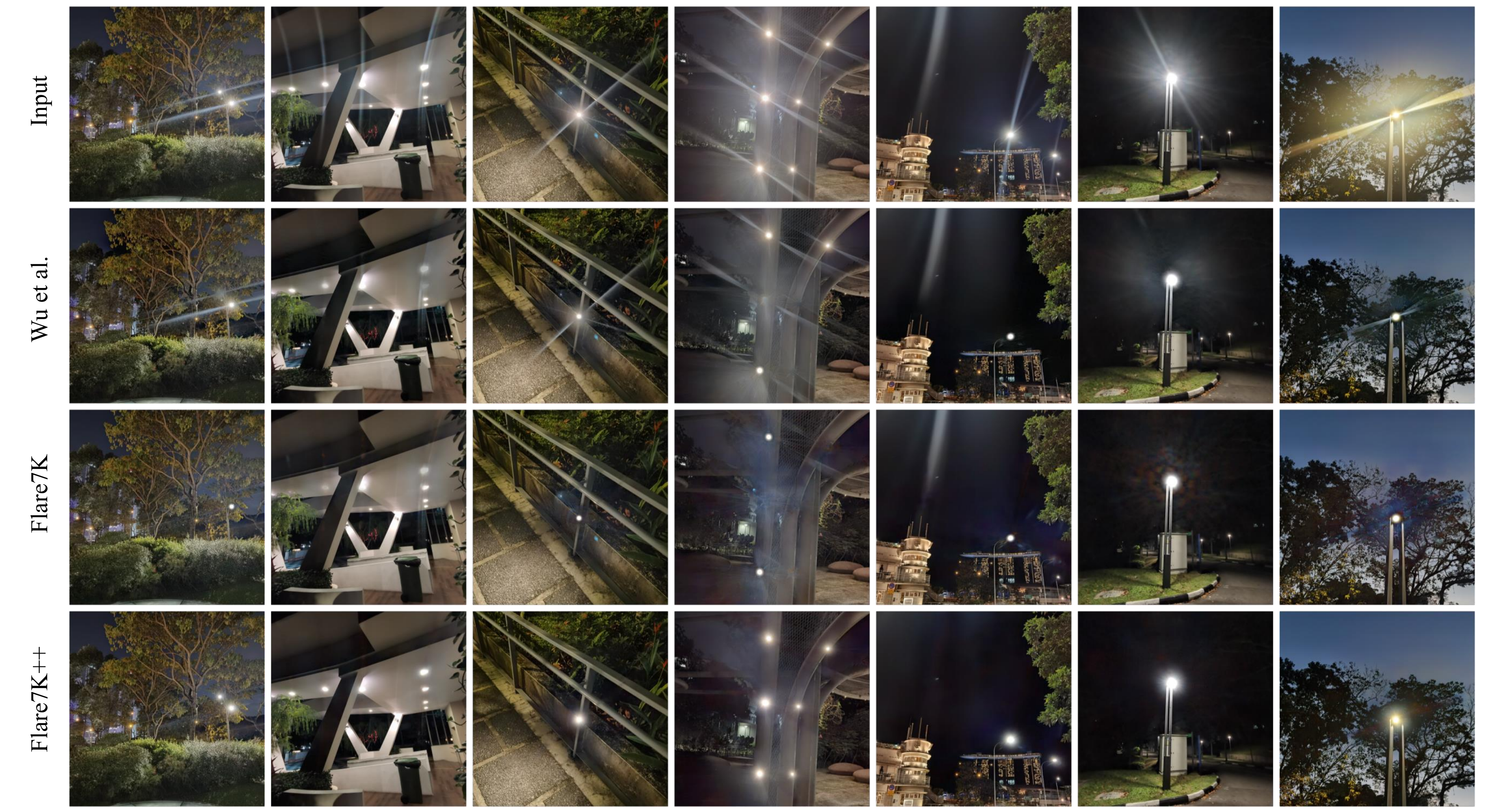}
	\caption{\pami{Comparison of different flare removal methods on real-world nighttime flare-corrupted images. Compared to results obtained only using Flare7K, our approach trained with Flare7K++ is able to significantly reduce severe image degradation around the light source.}}
	\label{fig:real_comparison}
	\vspace{-3mm}
\end{figure*}

\begin{figure*}
	\centering
	\includegraphics[width=1.0\linewidth]{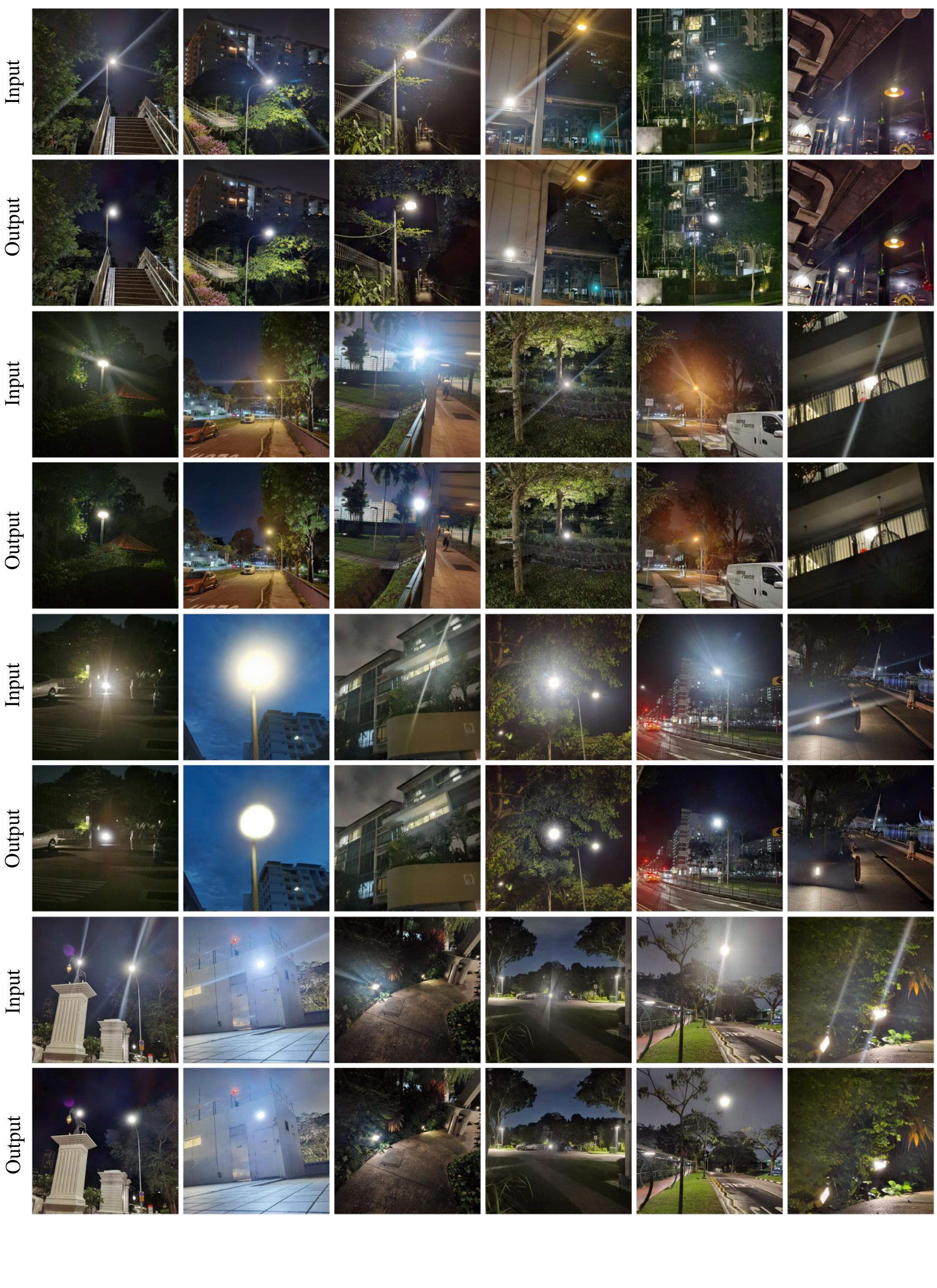}
 	\vspace{-20mm}
	\caption{\pami{Our results on real-world nighttime flare-corrupted images captured with different lenses. Since our Flare7K++ dataset contains diverse flares, it can generalize well to different situations.}}
	\label{fig:real_comparison2}
\end{figure*}

\pami{
\subsection{Experimental Details}

\textbf{Data Augmentation.} \label{data_augmentation}
To train our nighttime flare removal model, our paired flare-corrupted and flare-free images are generated on the fly.  The background images are sampled from the 24K Flickr images~\cite{reflectionk_zhang}.
Then, we sample the flare image and its corresponding light source from both Flare7K and Flare-R datasets with a probability of 50\% each.
An inverse gamma correction with $\gamma \sim U(1.8,2.2)$ is first applied to the flare image (including light source) and background image to recover the linear luminance.
For the image in our flare dataset, a random rotation $U(0,2\pi)$, a random translation $U(-300,300)$, a random shear $U(-\pi/9,\pi/9)$, a random scale $U(0.8,1.5)$, a random blur with the blur size in $U(0.1,3)$, and a random flip are applied to the paired light source and flare image. Then, a random global color offset in $U(-0.02,0.02)$ will be added to the flare image to simulate the situation in which the flare may illuminate the whole scene. 
For each background image, we randomly multiply the RGB values with $U(0.5,1.2)$ and add a Gaussian noise with variance sampled from a scaled chi-square distribution $\sigma^2 \sim 0.01\chi^2$ to it. After adding the augmented light source image to the background, the result is then clipped to $[0,1]$ as the corresponding flare-free image ground truth.
Then, we generate the ground truth for the flare by subtracting the light source from the augmented flare image as Fig.~\ref{fig:flare_removal_pipeline}(b).
Finally, we combine the augmented flare and background images to produce the flare-corrupted images, which serve as input to our network.

\noindent
\textbf{Training Details.}
During the training stage, our input flare-corrupted images are cropped to $512 \times 512 \times 3$ with a batch size of 2. We train our model for 300K iterations on the 24K Flickr image dataset~\cite{reflectionk_zhang} with the ADAM optimizer. The learning rate is set to $10^{-4}$.
To build a benchmark on our Flare7K++ dataset, we retrain the state-of-the-art image restoration networks including MPRNet~\cite{MPRNet}, HINet~\cite{HINet}, Uformer~\cite{Uformer}, and Restormer~\cite{Restormer} using our Flare7K++ dataset.
As MPRNet and HINet are multi-stage image restoration networks that are designed to handle input and output channel numbers that are consistent with each other, we add a 3-channel black image to the input. By doing so, we ensure that the number of channels in the input is equal to 6, which is necessary for the proper functioning of the networks.
To ensure the fairness of the experiment, we use the same training settings and data augmentation methods as stated above to train these methods. 
All these models are trained using the Nvidia Geforce RTX 3090 GPUs.
Due to the limited GPU memory (24G memory) of Nvidia Geforce RTX 3090, we reduce the parameters of the MPRNet and Restormer which are known as the heavy networks.
Specifically, we decrease the refinement block number of Restormer from 4 to 1 and set the dimension of the feature channel to 16 rather than the default dimension 48.
The MPRNet's number of features is set to 24 rather than the default 40 to satisfy the memory limitation.
We believe the default parameter settings of these two networks would perform better.
}

\subsection{Comparison with Previous Work}
To demonstrate the effectiveness and advantages of our dataset, we compare the performance of different datasets and methods for nighttime flare removal.
We also present a benchmark of the existing image restoration methods on our dataset.

\noindent
\textbf{Experimental Setting.}
In the absence of nighttime flare removal methods, we compare our baseline model with a nighttime dehazing method~\cite{nighttime_zhang}, a nighttime visual enhancement method~\cite{nighttime_sharma}, and a flare removal method~\cite{how_to}. 
Zhang~\textit{et~al.}~\cite{nighttime_zhang} propose a nighttime haze synthesis method that can simulate light rays and train a network with the synthetic data.
We use its pre-trained model for comparison.
Sharma~\textit{et~al.}'s nighttime visual enhancement method~\cite{nighttime_sharma} is an unsupervised test-time training method that can extract low-frequency light effects based on gray world assumption and glare-smooth prior.  
We follow its setting to test the results on our test data.
Wu~\textit{et~al.}'s study~\cite{how_to} is most related to our work. Since Wu~\textit{et~al.} do not provide their model checkpoint, we use their released code and data to train a model for comparison. 
\pami{We also compare our method with previous Flare7K baseline that uses Uformer~\cite{Uformer} as backbone. 
}
The quantitative results are presented in Table \ref{tab:PSNR_SSIM}.

\noindent
\textbf{Qualitative Comparison.} 
We first show the visual comparison of real-world nighttime flares in Fig~\ref{fig:result_comparison}. 
The comparison suggests that recent approaches for nighttime haze removal~\cite{nighttime_zhang} and nighttime visual enhancement~\cite{nighttime_sharma} have little effect on nighttime lens flare. In contrast, the model trained on our dataset can produce satisfactory outputs on nighttime flare-corrupted images.
\pami{
Although Wu~\textit{et~al.}~\cite{how_to} can eliminate the glare effect effectively, it cannot tackle diverse streak effects and may even remove the light source.}
We show more results of our method in Fig \ref{fig:real_comparison}. In these challenging cases, our approach yields satisfactory flare-free results.
\pami{With the help of the Flare-R dataset, our new method can eliminate complicated degradation around the light source better.
Besides, our end-to-end pipeline can solve the problem that tiny and not overexposed light sources are hard to recover.
Our method can even achieve robust results in scenarios with multiple light sources.
}

\noindent
\textbf{Quantitative Comparison.} We use common full-reference metrics PSNR, SSIM~\cite{ssim}, and LPIPS~\cite{lpips} to quantify the performance of different methods in Table \ref{tab:PSNR_SSIM}.
\pami{
To better illustrate the comparison of different methods in terms of their effectiveness in removing different components of flares, we also introduce two metrics, S-PSNR and G-PSNR, to measure the flare removal performance on glare and streak's region.
The mask of glare and streak of test data is manually plotted as shown in Fig.~\ref{fig:test_type}.}
Since Sharma~\textit{et~al.}'s method~\cite{nighttime_sharma} is based on the gray world assumption, it fails in the glare effect in white.
Zhang~\textit{et~al.}'s method~\cite{nighttime_zhang} is mainly designed for nighttime haze.
Although it can alleviate the lens flare, it also changes the image's color, leading to a decrease in SSIM.
\pami{
In Table \ref{tab:PSNR_SSIM}, the difference between U-Net~\cite{unet} on our dataset and Wu~\textit{et~al.}'s method also suggests the effectiveness of Flare7K++.
In comparison, Uformer \cite{Uformer} performs best for PSNR and LPIPS.
Thus, we follow the Flare7K's setting and set Uformer~\cite{Uformer} as our baseline method in the following sections.}
All these networks achieve good performance, revealing the reliability of our dataset. 
\pami{More results of our method are shown in Fig.~\ref{fig:real_comparison2}.}

\pami{
\subsection{Ablation Study}
In this section, we study the main designs of our proposed method and detailed comparison with the previous Flare7K dataset.

\begin{figure}[t]
  \centering
   \includegraphics[width=1.0\linewidth]{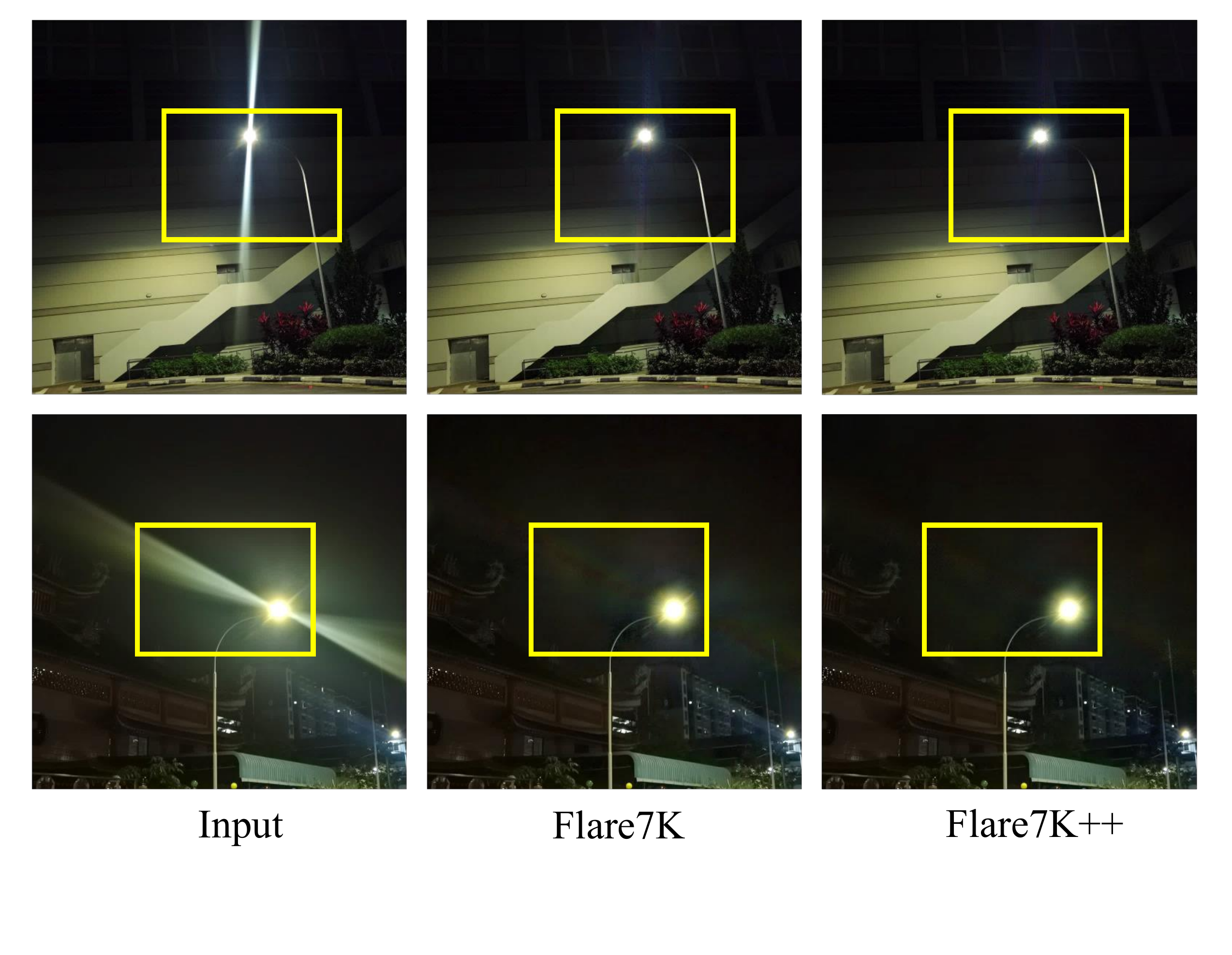}
   \vspace{-14mm}
   \caption{\pami{Visual comparison for training networks with Flare7K and Flare7K++. Bright light sources may leave complicated artifacts on real-captured nighttime scenes. The network only training with Flare7K cannot solve the degradation well. Training with the combination of Flare-R and Flare7K can solve these problems effectively.} }
   \label{fig:ablation_dataset}
\end{figure}

\begin{table}[th]
\caption{\pami{Ablation study of training network with different subsets of Flare7K++.}}
\label{tab:dataset_ablation}
\resizebox{0.48\textwidth}{!}{
\begin{tabular}{cclccccc}
\toprule
\multicolumn{2}{c}{Training sets} &  & \multicolumn{5}{c}{Real captured test set} \\ \cmidrule{1-2} \cmidrule{4-8} 
Flare7K         & Flare-R         &  & PSNR    & SSIM  & LPIPS  & G-PSNR & S-PSNR \\ \midrule
$\checkmark$    &                 &  & 27.257  & 0.890 & 0.0471 & 23.762 & 21.294 \\ \cmidrule{4-8} 
$\checkmark$    & $\checkmark$    &  & \textbf{27.633}  & \textbf{0.894} & \textbf{0.0428} & \textbf{23.949} & \textbf{22.603} \\ \midrule
\end{tabular}}
\end{table}

\noindent
\textbf{Dataset.}
To illustrate the necessity of training with both Flare-R and Flare7K, we conduct an ablation study using only Flare7K in our new pipeline.
The results, shown in Fig.~\ref{fig:ablation_dataset}, reveal that areas near the light source often exhibit severe artifacts. 
As these artifacts are difficult to synthesize, training the network solely with Flare7K may not fully remove them from the image. 
However, by incorporating Flare-R into the training process, our mix-training strategy greatly improves flare removal performance in streak regions and effectively resolves the artifact issue, as demonstrated in Table.~\ref{tab:dataset_ablation}.

\noindent
\textbf{Light Source Completion.}
To demonstrate the effectiveness of our light source processing method, we conduct an ablation study to evaluate the impact of incorporating light source annotations during training. 
For comparison, we train a Uformer~\cite{Uformer} with 6-channel output without using light source annotations as shown in Fig~\ref{fig:flare_removal_pipeline}(b).
Then, we follow Flare7K's setting to set a mask at the input's overexposed regions and not supervise the output in these regions by using masked flare loss and masked background loss.
In the inference stage, we set the threshold of luminance to 0.97 to extract a mask and use Wu~\textit{et~al.}~\cite{how_to}'s mask feathering method to smoothly blend the light sources back to the image. 
As shown in Table~\ref{tab:light_ablation}, training without light source annotations may seriously decrease the quantitative results. Besides, Fig.~\ref{fig:ablation_light} illustrates the benefits of our new light source annotations. It can help reconstruct tiny light sources well and can avoid leaving a tail of the streak effect next to the light source.

\begin{figure}[t]
  \centering
   \includegraphics[width=1.0\linewidth]{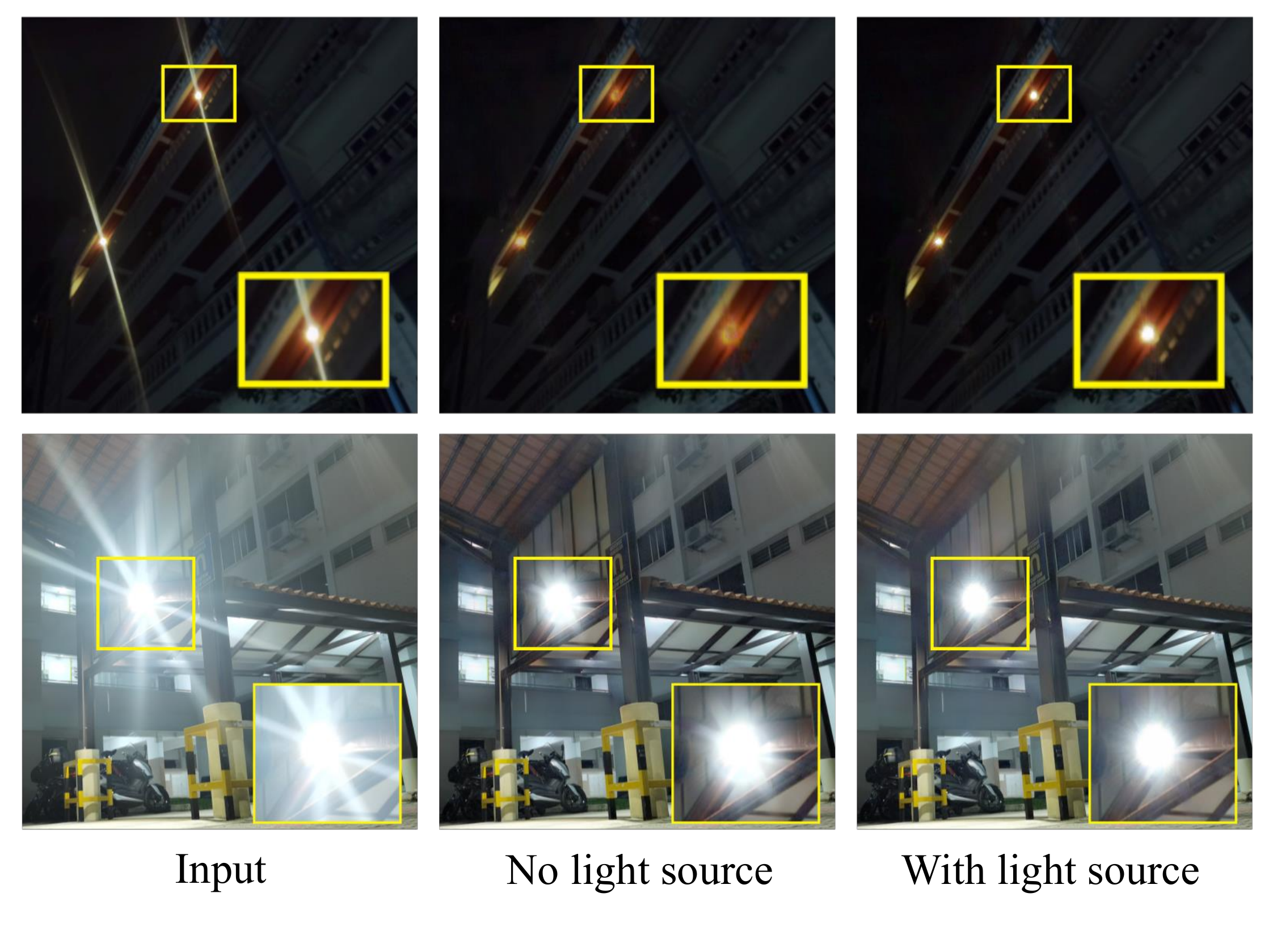}
   \caption{\pami{Visual comparison for training networks with and without light source annotations. The mask feathering method may also remove tiny light sources or preserve parts of the streak around the light source. Our light source annotations can help solve these problems.} }
   \label{fig:ablation_light}
\end{figure}
}

\begin{figure}[t]
  \centering
   \includegraphics[width=1.0\linewidth]{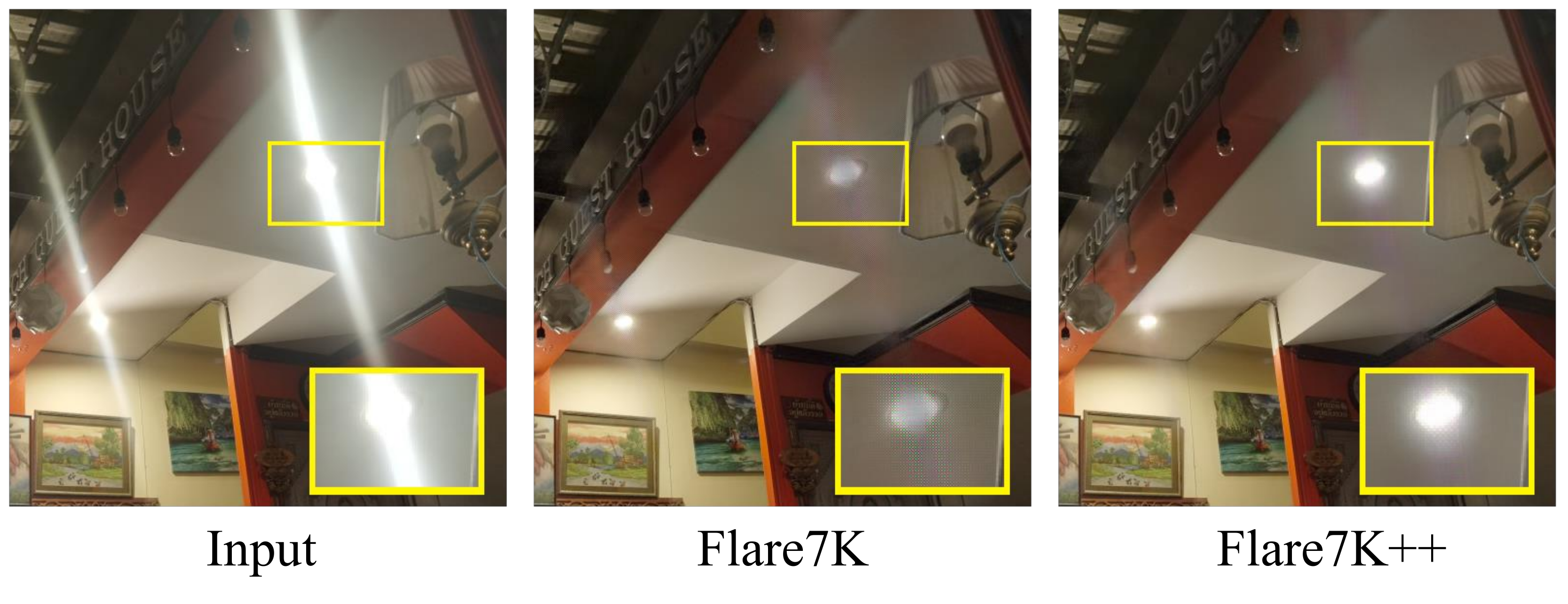}
   \caption{\pami{Visual comparison for training networks with Flare7K's pipeline and our new pipeline. While training the network, we do not change the loss function. All networks are trained with Flare7K++.} }
   \vspace{-3mm}
   \label{fig:ablation_pipeline}
\end{figure}

\begin{table}[th]
\caption{\pami{Ablation study of training with and w/o light source annotations.}}
\label{tab:light_ablation}
\resizebox{0.48\textwidth}{!}{
\begin{tabular}{cllccccc}
\toprule
\multicolumn{2}{c}{Light source}      &  & PSNR   & SSIM  & LPIPS  & G-PSNR & S-PSNR \\ \midrule
\multicolumn{2}{c}{w/o light source}  &  & 26.850 & \textbf{0.895} & 0.0473 & 23.441 & 21.909 \\ \midrule
\multicolumn{2}{c}{with light source} &  & \textbf{27.633} & 0.894 & \textbf{0.0428} & \textbf{23.949} & \textbf{22.603} \\ \midrule
\end{tabular}}
\end{table}

\begin{figure*}[t]
  \centering
   \includegraphics[width=1.0\linewidth]{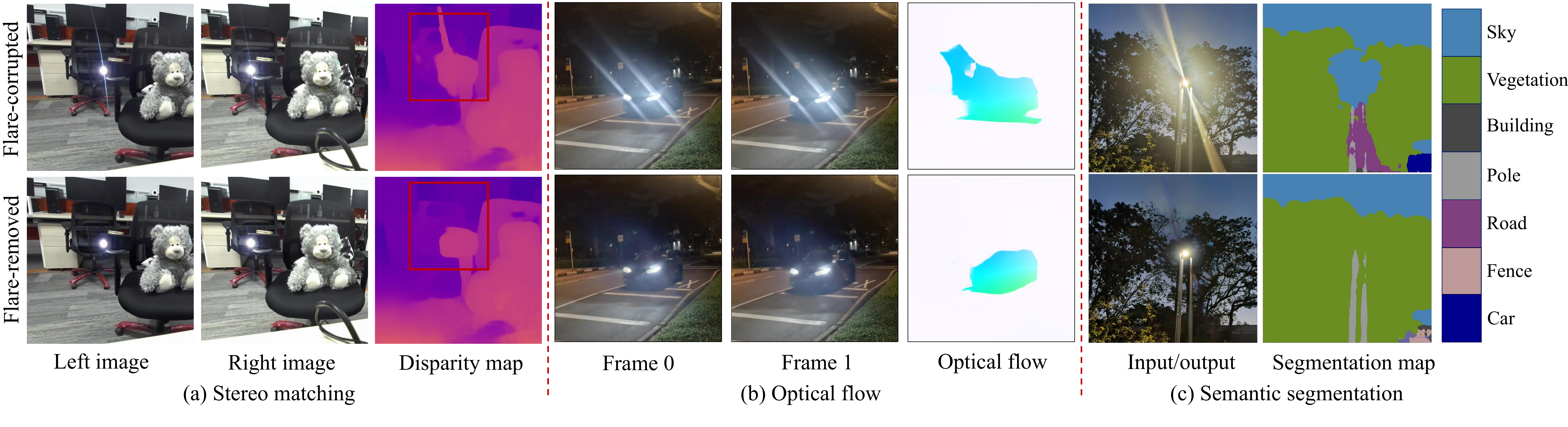}
   \caption{\pami{Visual comparison of the effectiveness of flare removal algorithms on various downstream tasks, including stereo matching, semantic segmentation, and optical flow. In~(a), our flare removal algorithm prevents mismatching caused by lens flare, improving the accuracy of stereo matching. In~(b), lens flare is incorrectly detected as a part of the car, causing errors in the optical flow estimation algorithm. Removing the flare improves the accuracy of the estimation. In~(c), the presence of lens flare leads to the misclassification of objects in the scene, resulting in inaccurate semantic segmentation. By removing the lens flare, more accurate semantic segmentation results can be obtained.} }
   \label{fig:downstream_all}
\end{figure*}

\pami{
\noindent
\textbf{Training Pipeline.}
In Wu~\textit{et~al.} and Flare7K's training pipeline~\cite{how_to}, the network will output a flare-removed image. 
Then, the difference between the flare-corrupted image and the flare-removed image will be calculated to estimate the flare.
To avoid the influence brought by the light source annotations, we train a network with the pipeline of Fig~\ref{fig:flare_removal_pipeline}(a) and also supervise the overexposed region of the estimated flare-removed and flare image by not adopting masked loss.
In the saturated region of the flare-corrupted image, the addition of flare-free ground truth and ground truth of flare image will get higher than 1.0.
Thus, the difference operation will introduce bias while calculating the flare image during training.
This bias may bring artifacts at streak's overexposed region as shown in Fig~\ref{fig:ablation_pipeline}.
These artifacts will influence the quantitative results seriously as shown in Table.~\ref{tab:pipeline_ablation}.
Since our new pipeline does not contain the subtraction operation, it can avoid this problem and reconstruct details well in the saturated regions.

\begin{table}[th]
\caption{\pami{Ablation study of training with our proposed pipeline.}}
\label{tab:pipeline_ablation}
\resizebox{0.48\textwidth}{!}{
\begin{tabular}{llccccc}
\toprule
Pipeline          &  & \multicolumn{1}{l}{PSNR} & SSIM           & LPIPS           & G-PSNR          & \multicolumn{1}{l}{S-PSNR} \\ \midrule
Wu~\textit{et~al.}~\cite{how_to}          &  & 27.166                   & 0.861          & 0.0432          & 23.598          & 22.118                     \\ \midrule
Ours   &  & \textbf{27.633}          & \textbf{0.894} & \textbf{0.0428} & \textbf{23.949} & \textbf{22.603}            \\ \midrule
\end{tabular}}
\end{table}

\subsection{Flare Removal for Downstream Tasks}
To show the effectiveness of the flare removal algorithm trained on our Flare7K++ for downstream tasks, we evaluate the performance of both the originally captured images and the processed images in common nighttime automatic driving tasks such as stereo matching, semantic segmentation, and optical flow estimation as shown in Fig~\ref{fig:downstream_all}.

\noindent
\textbf{Stereo Matching.}
In this task, the apparent pixel difference for each corresponding pixel in stereo images is calculated to obtain the disparity map. 
As shown in Fig.~\ref{fig:downstream_all}(a), the stereo images in the first row are captured by the ZED camera, and the second-row images are flare-removed images using a deep model trained on our Flare7K++ dataset. 
Then, LEAStereo~\cite{LEAStereo}, a state-of-the-art stereo matching algorithm pre-trained on the Kitti dataset~\cite{kitti}, is applied to these stereo images to estimate the disparity.
The red boxes indicate the obvious differences.
Our results demonstrate that our algorithm effectively avoids mismatching caused by lens flare, significantly improving the robustness of the stereo matching algorithm.

\noindent
\textbf{Optical Flow.}
Our flare removal method also benefits optical flow estimation, as demonstrated in Fig.\ref{fig:downstream_all}(b). 
The optical flow was estimated using RAFT~\cite{teed2020raft} trained on Sintel~\cite{Sintel} dataset. 
The figure shows that the flare caused by car light leads to erroneous optical flow.
Our method effectively eliminates the flare, resulting in more precise and dependable optical flow estimation. 
This leads to an overall improvement in the performance and safety of downstream tasks relying on optical flow.

\noindent
\textbf{Semantic Segmentation.}
Fig.~\ref{fig:downstream_all}(c) illustrates the benefit of our flare removal method for image segmentation.
In the figure, the segmentation maps are calculated by DANNet~\cite{DANNet,DANNet_pami}, a nighttime semantic segmentation algorithm trained on Dark Zurich~\cite{darkzurich} and Cityscapes~\cite{cordts2016cityscapes}.
In the flare-corrupted image, the vegetation is mislabeled as the sky and the pole is mislabeled as the road, which may pose potential risks for nighttime driving. 
Our flare removal method effectively addresses these issues by improving the accuracy and reliability of the segmentation maps.

}

\begin{figure} [t]
	\centering
	\includegraphics[width=1.0\linewidth]{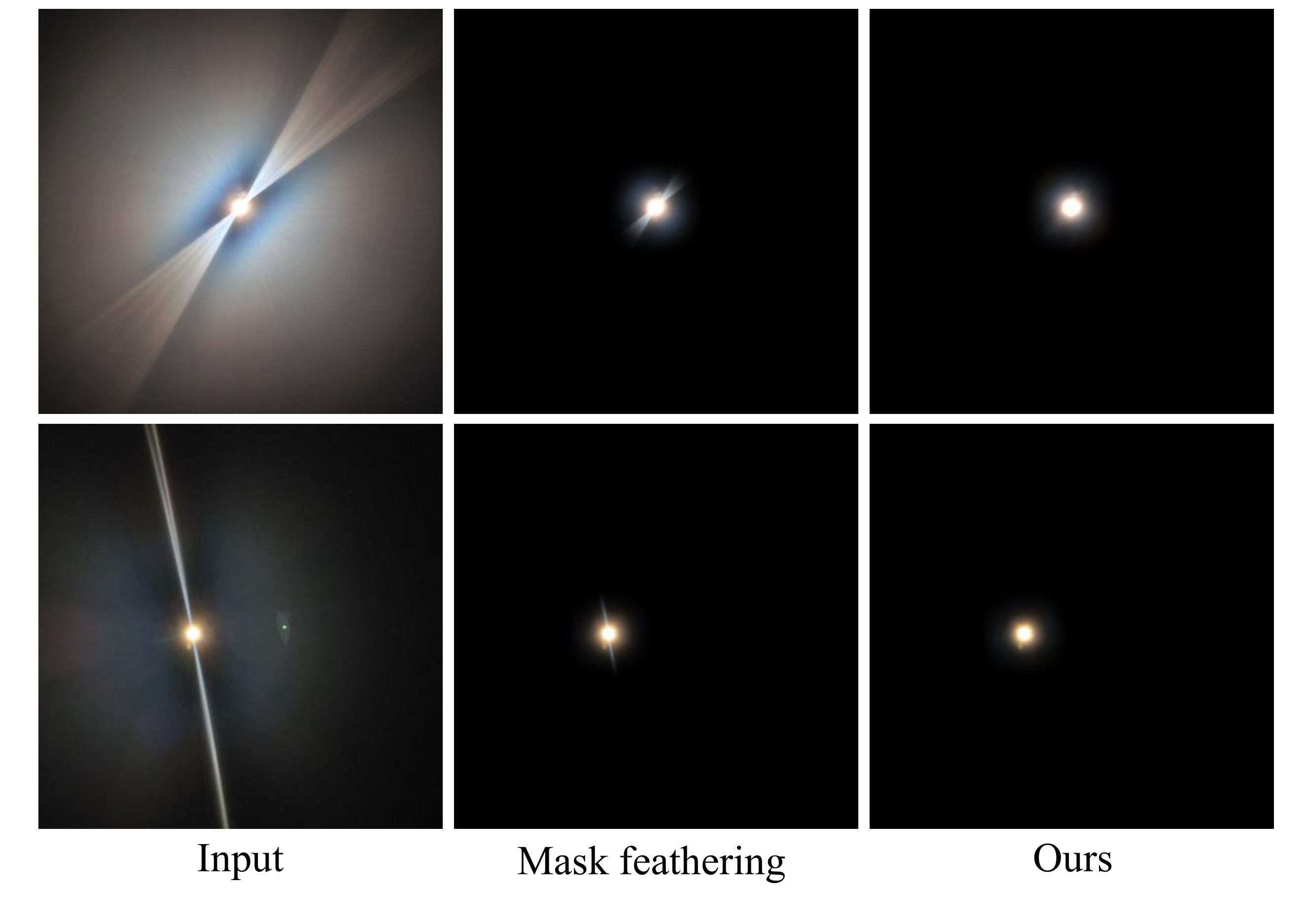}
	\caption{\pami{Light source extraction with our dataset. Compared with traditional threshold and mask feathering method, our light source extraction method can avoid leaving a little "tail". }} 
	\vspace{-3mm}
	\label{fig:light_extraction}
\end{figure}

\pami{
\section{Potential applications}\label{sec:extended_applications}

In our dataset, for each lens flare image, we provide separated images of light source, glare with shimmer, streak, and reflective flare. These annotations can facilitate the design of improved flare removal methods and promote other related tasks. 
Besides, as the first nighttime flare dataset, the flare images can also be added to the training dataset of other nighttime vision algorithms to increase the robustness for flare-corrupted situations.
With these annotations of our data, we provide more details about how to implement lens flare segmentation and light source extraction as follows.
Please note that these annotations are not limited to these two applications.

\noindent
\textbf{Lens Flare Segmentation.}
Flare segmentation is useful for flare removal.
When the streak is too bright or overexposed, it is difficult to recover the details of these regions by just using an image decomposition network. A streak segmentation model trained on flare segmentation data can help locate these regions. Then, an image inpainting algorithm can be used to restore the missing information.
Besides, the reflective flares for smartphone lenses always have the same patterns as the light sources' brightest regions.
For matrix LED lights, the reflective flares will also be matrix-shaped patterns as shown in Fig.~\ref{fig:reflective_flare_generation} of the main paper.
Thus, segmented reflective flares can be referred to restore the brightest regions in the saturated area, achieving the HDR snapshot reconstruction.

To demonstrate the effectiveness of flare segmentation, we train a network for lens flare segmentation and show the flare segmentation results in Fig.~\ref{fig:streak_segmentation}. 
Specifically, we use the PSPNet\cite{pspnet} as the flare segmentation network and train it using the flare annotation of our dataset for 80k iterations with a batch size of 2 on an Nvidia GTX 1080 GPU. The data augmentation pipeline is the same as the pipeline mentioned in Section \ref{data_augmentation}. 
We use the SGD optimizer with a learning rate $0.01$, a weight decay $0.0005$, and a poly learning rate policy. 
The power rate of the policy is set to $0.9$. Since areas of the streak and light source are remarkably lower than the glare effect, we use a cross-entropy loss with the class weights 1.0, 1.0, 2.0, and 4.0 for background, glare with shimmer, streak, and light source.

\begin{figure} [t]
	\centering
	\includegraphics[width=0.8\linewidth]{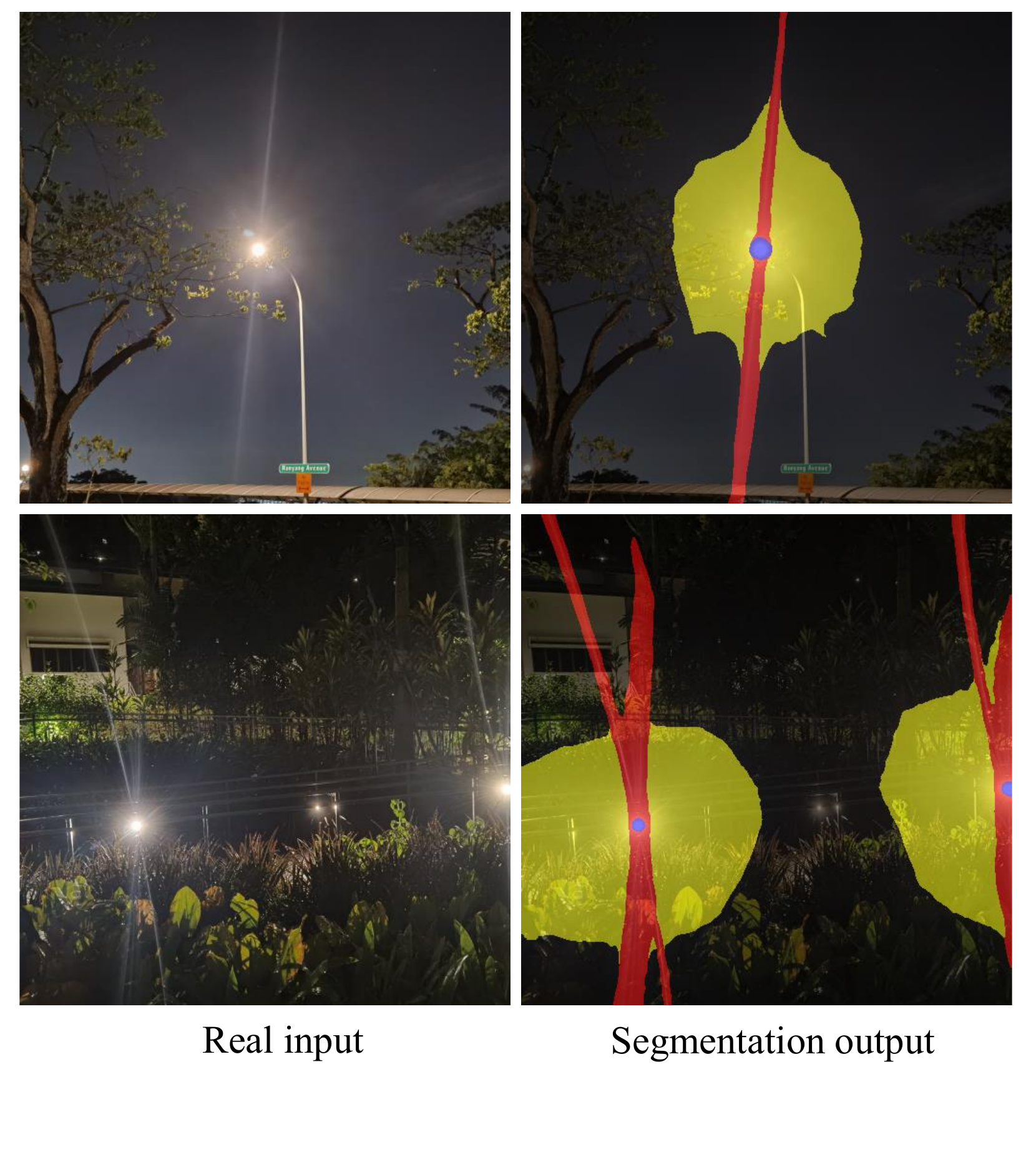}
	\vspace{-7mm}
	\caption{Lens flare segmentation with our dataset. Since our dataset provides separated images for each component, it can be used for lens flare segmentation. This figure shows that PSPNet~\cite{pspnet} trained on our dataset with annotations can accurately segment different components in real-world flare-corrupted images. In this figure, red, yellow, and blue represent streak, glare, and light source, respectively.} 
	\vspace{-4mm}
	\label{fig:streak_segmentation}
\end{figure}

\noindent
\textbf{Light Source Extraction.} \label{light_source_extraction}
Extracting light sources from a flare image has always been a challenging task. 
If only overexposed areas in the image are extracted as light sources, the final result will be very unrealistic. 
Therefore, retaining a certain degree of glare effect is the key to achieving high-quality light source reconstruction. 
However, even for flare images taken in a darkroom, it is difficult to reconstruct the light source using traditional algorithms.
As shown in Fig.~\ref{fig:light_extraction}, the method of threshold and mask feathering~\cite{how_to} will also lead to preserving parts of streaks in the image.
Even worse, when the streak gets overexposed, it may be all extracted as a light source.

With the light source annotations provided by our Flare7K dataset, paired scattering flare and light source can be used to train a light source extraction network. 
We use a U-Net as the light source extraction network and train it with $L_1$ loss on these 5000 scattering flare images for 50k iterations (20 epochs) on an Nvidia GTX 3090 GPU. 
Like our flare removal model, the batch size is set to 2. Moreover, we use the Adam optimizer with a learning rate $10^{-4}$.
As shown in Fig.~\ref{fig:light_extraction}, the glare effect of our method around the light source is more natural when compared with Wu et al.~\cite{how_to}'s mask feathering method. It benefits from the high-quality annotations of the light source images in our scattering flares.
The light source annotations of our Flare-R dataset are also obtained from our trained light source extraction method.

}

\vspace{-3mm}
\section{Limitations}
\pami{
The aforementioned experiments demonstrate the impressive performance of Flare7K++ in various scenarios
However, the networks trained on our data may still fail to address some challenging cases. 
Since reflective flares always produce bright spots similar to bright windows and street lights in the distance at night, these windows or lights may also be removed as a part of reflective flares.
Besides, when the light source is close to the camera, it may leave a large glare that may cover the whole image.
This kind of glare is tough to be removed. 
In this situation, the streak and the region around the light source may get saturated in one or more channels. 
Existing image decomposition-based methods cannot complete this missing information well. 
These limitations are mainly caused by the flare removal method rather than our dataset. 
To solve these problems, one would need to consider semantic priors as input or introduce better network structures.} 

\vspace{-3mm}
\section{Conclusion}
\label{conclusion}
\pami{
In this paper, we present a new dataset, Flare7K++, aiming at advancing nighttime flare removal. 
Flare7K++ is composed of a real-captured flare image dataset Flare-R and a synthetic flare image dataset Flare7K.
Flare7K contains 5,000 scattering and 2,000 reflective flare images, consisting of 25 types of scattering flares and 10 types of reflective flares.
Flare-R contains 962 flare patterns and is captured in the darkroom with different types of lens contaminants.
Besides, we provide rich annotations for different flare components, like light source, streak, glare with shimmer, which are always absent in the previous dataset.
With the proposed dataset, we design a new nighttime flare removal pipeline to improve the quality of nighttime images and boost the stability of nighttime vision algorithms. 
Our novel pipeline efficiently leverages the light source annotations, resulting in improved accuracy in retaining the light source while eliminating the flare.
Extensive experiments show that our dataset is sufficient to facilitate the removal of different types of nighttime lens flares. 
}

\ifCLASSOPTIONcaptionsoff
  \newpage
\fi



%

\medskip
{
\small
\bibliographystyle{plain}
\bibliography{reference}

\begin{thebibliography}{10}

\bibitem{auto_removal}
C~S Asha, Sooraj~Kumar Bhat, Deepa Nayak, and Chaithra Bhat.
\newblock Auto removal of bright spot from images captured against flashing
  light source.
\newblock In {\em IEEE International Conference on Distributed Computing, VLSI,
  Electrical Circuits and Robotics}, 2019.

\bibitem{Sintel}
D.~J. Butler, J.~Wulff, G.~B. Stanley, and M.~J. Black.
\newblock A naturalistic open source movie for optical flow evaluation.
\newblock In {\em European Conference on Computer Vision}, 2012.

\bibitem{automated_removal}
Floris Chabert.
\newblock Automated lens flare removal, 2014.
\newblock Technical Report, Department of Electrical Engineering, Stanford
  University.

\bibitem{HINet}
Liangyu Chen, Xin Lu, Jie Zhang, Xiaojie Chu, and Chengpeng Chen.
\newblock Hinet: Half instance normalization network for image restoration.
\newblock In {\em IEEE Conference on Computer Vision and Pattern Recognition
  Workshops}, 2021.

\bibitem{LEAStereo}
Xuelian Cheng, Yiran Zhong, Mehrtash Harandi, Yuchao Dai, Xiaojun Chang,
  Hongdong Li, Tom Drummond, and Zongyuan Ge.
\newblock Hierarchical neural architecture search for deep stereo matching.
\newblock {\em Advances in Neural Information Processing Systems},
  33:22158--22169, 2020.

\bibitem{cordts2016cityscapes}
Marius Cordts, Mohamed Omran, Sebastian Ramos, Timo Rehfeld, Markus Enzweiler,
  Rodrigo Benenson, Uwe Franke, Stefan Roth, and Bernt Schiele.
\newblock The cityscapes dataset for semantic urban scene understanding.
\newblock In {\em IEEE Conference on Computer Vision and Pattern Recognition},
  2016.

\bibitem{dai2022flare7k}
Yuekun Dai, Chongyi Li, Shangchen Zhou, Ruicheng Feng, and Chen~Change Loy.
\newblock Flare7{K}: A phenomenological nighttime flare removal dataset.
\newblock In {\em Thirty-sixth Conference on Neural Information Processing
  Systems Datasets and Benchmarks Track}, 2022.

\bibitem{dai2023nighttime}
Yuekun Dai, Yihang Luo, Shangchen Zhou, Chongyi Li, and Chen~Change Loy.
\newblock Nighttime smartphone reflective flare removal using optical center
  symmetry prior.
\newblock In {\em IEEE Conference on Computer Vision and Pattern Recognition},
  2023.

\bibitem{Ernst2005}
Manfred Ernst, Tomas Akenine-Moller, and Henrik~Wann Jensen.
\newblock Interactive rendering of caustics using interpolated warped volumes.
\newblock In {\em Graphics Interface}, 2005.

\bibitem{feng2023generating}
Ruicheng Feng, Chongyi Li, Huaijin Chen, Shuai Li, Jinwei Gu, and Chen~Change
  Loy.
\newblock Generating aligned pseudo-supervision from non-aligned data for image
  restoration in under-display camera.
\newblock In {\em IEEE Conference on Computer Vision and Pattern Recognition},
  2023.

\bibitem{feng2021removing}
Ruicheng Feng, Chongyi Li, Huaijin Chen, Shuai Li, Chen~Change Loy, and Jinwei
  Gu.
\newblock Removing diffraction image artifacts in under-display camera via
  dynamic skip connection network.
\newblock In {\em IEEE Conference on Computer Vision and Pattern Recognition},
  2021.

\bibitem{kitti}
Andreas Geiger, Philip Lenz, Christoph Stiller, and Raquel Urtasun.
\newblock Vision meets robotics: The kitti dataset.
\newblock {\em The International Journal of Robotics Research},
  32(11):1231--1237, 2013.

\bibitem{gu2009removing}
Jinwei Gu, Ravi Ramamoorthi, Peter Belhumeur, and Shree Nayar.
\newblock Removing image artifacts due to dirty camera lenses and thin
  occluders.
\newblock In {\em ACM SIGGRAPH Asia}, 2009.

\bibitem{dehaze_he}
Kaiming He, Jian Sun, and Xiaoou Tang.
\newblock Single image haze removal using dark channel prior.
\newblock {\em IEEE Transactions on Pattern Analysis and Machine Intelligence},
  33(12):2341--2353, 2010.

\bibitem{holladay1926fundamentals}
Lloyd~L Holladay.
\newblock The fundamentals of glare and visibility.
\newblock {\em JOSA}, 12(4):271--319, 1926.

\bibitem{flare_simulation1}
Matthias Hullin, Elmar Eisemann, Hans-Peter Seidel, and Sungkil Lee.
\newblock Physically-based real-time lens flare rendering.
\newblock In {\em ACM SIGGRAPH}, 2011.

\bibitem{flare_simulation2}
Sungkil Lee and Elmar Eisemann.
\newblock Practical real-time lens-flare rendering.
\newblock {\em Computer Graphics Forum}, 32(4):1--6, 2013.

\bibitem{li2021let}
Xiaoyu Li, Bo~Zhang, Jing Liao, and Pedro~V Sander.
\newblock Let's see clearly: Contaminant artifact removal for moving cameras.
\newblock In {\em IEEE International Conference on Computer Vision}, 2021.

\bibitem{li_nighttime_2015}
Yu~Li, Robby~T. Tan, and Michael~S. Brown.
\newblock Nighttime haze removal with glow and multiple light colors.
\newblock In {\em IEEE International Conference on Computer Vision}, 2015.

\bibitem{shedding_2003}
Srinivasa~G. Narasimhan and Shree~K. Nayar.
\newblock Shedding light on the weather.
\newblock In {\em IEEE Conference on Computer Vision and Pattern Recognition},
  2003.

\bibitem{nightowls}
Lukáš Neumann, Michelle Karg, Shanshan Zhang, Christian Scharfenberger, Eric
  Piegert, Sarah Mistr, Olga Prokofyeva, Robert Thiel, Andrea Vedaldi, Andrew
  Zisserman, and Bernt Schiele.
\newblock Nightowls: A pedestrians at night dataset.
\newblock In {\em Asian Conference on Computer Vision}, 2018.

\bibitem{aerial_tracking}
Andreas Nussberger, Helmut Grabner, and Luc~Van Gool.
\newblock Robust aerial object tracking in images with lens flare.
\newblock In {\em IEEE International Conference on Robotics and Automation},
  2015.

\bibitem{light_source}
Xiaotian Qiao, Gerhard~P. Hancke, and Rynson W.~H. Lau.
\newblock Light source guided single-image flare removal from unpaired data.
\newblock In {\em IEEE International Conference on Computer Vision}, 2021.

\bibitem{HDR_1}
Erik Reinhard, Wolfgang Heidrich, Paul Debevec, Sumanta Pattanaik, Greg Ward,
  and Karol Myszkowski.
\newblock High dynamic range imaging: Acquisition, display, and image-based
  lighting, 2010.

\bibitem{unet}
Olaf Ronneberger, Philipp Fischer, and Thomas Brox.
\newblock U-net: Convolutional networks for biomedical image segmentation.
\newblock In {\em International Conference on Medical Image Computing and
  Computer-Assisted Intervention}, 2015.

\bibitem{rouf2011glare}
Mushfiqur Rouf, Rafa{\l} Mantiuk, Wolfgang Heidrich, Matthew Trentacoste, and
  Cheryl Lau.
\newblock Glare encoding of high dynamic range images.
\newblock In {\em IEEE Conference on Computer Vision and Pattern Recognition},
  2011.

\bibitem{darkzurich}
Christos Sakaridis, Dengxin Dai, and Luc~Van Gool.
\newblock Map-guided curriculum domain adaptation and uncertainty-aware
  evaluation for semantic nighttime image segmentation.
\newblock {\em IEEE Transactions on Pattern Analysis and Machine Intelligence},
  44(6):3139--3153, 2020.

\bibitem{nighttime_sharma}
Aashish Sharma and Robby~T. Tan.
\newblock Nighttime visibility enhancement by increasing the dynamic range and
  suppression of light effects.
\newblock In {\em IEEE Conference on Computer Vision and Pattern Recognition},
  2021.

\bibitem{rank_1}
Qilin Sun, Ethan Tseng, Qiang Fu, Wolfgang Heidrich, and Felix Heide.
\newblock Learning rank-1 diffractive optics for single-shot high dynamic range
  imaging.
\newblock In {\em IEEE Conference on Computer Vision and Pattern Recognition},
  2020.

\bibitem{HDR_2}
Eino-Ville Talvala, Andrew Adams, Mark Horowitz, and Marc Levoy.
\newblock Veiling glare in high dynamic range imaging.
\newblock {\em ACM Transactions on Graphics}, 26(3):37–es, 2007.

\bibitem{teed2020raft}
Zachary Teed and Jia Deng.
\newblock Raft: Recurrent all-pairs field transforms for optical flow.
\newblock In {\em European Conference on Computer Vision}, 2020.

\bibitem{auto_removal2}
Patricia Vitoria and Coloma Ballester.
\newblock Automatic flare spot artifact detection and removal in photographs.
\newblock {\em Journal of Mathematical Imaging and Vision}, 61:515--533, 2019.

\bibitem{Uformer}
Zhendong Wang, Xiaodong Cun, Jianmin Bao, Wengang Zhou, Jianzhuang Liu, and
  Houqiang Li.
\newblock Uformer: A general u-shaped transformer for image restorationn.
\newblock In {\em IEEE Conference on Computer Vision and Pattern Recognition},
  2022.

\bibitem{ssim}
Zhou Wang, Alan~C Bovik, Hamid~R Sheikh, and Eero~P Simoncelli.
\newblock Image quality assessment: from error visibility to structural
  similarity.
\newblock {\em IEEE transactions on image processing}, 13(4):600--612, 2004.

\bibitem{DANNet}
Xinyi Wu, Zhenyao Wu, Hao Guo, Lili Ju, and Song Wang.
\newblock {DANN}et: A one-stage domain adaptation network for unsupervised
  nighttime semantic segmentation.
\newblock In {\em IEEE Conference on Computer Vision and Pattern Recognition},
  2021.

\bibitem{DANNet_pami}
Xinyi Wu, Zhenyao Wu, Lili Ju, and Song Wang.
\newblock A one-stage domain adaptation network with image alignment for
  unsupervised nighttime semantic segmentation.
\newblock {\em IEEE Transactions on Pattern Analysis and Machine Intelligence},
  45(1):58--72, 2021.

\bibitem{how_to}
Yicheng Wu, Qiurui He, Tianfan Xue, Rahul Garg, Jiawen Chen, Ashok
  Veeraraghavan, and Jonathan~T. Barron.
\newblock How to train neural networks for flare removal.
\newblock In {\em IEEE International Conference on Computer Vision}, 2021.

\bibitem{nighttime_yan}
Wending Yan, Robby~T. Tan, and Dengxin Dai.
\newblock Nighttime defogging using high-low frequency decomposition and
  grayscale-color networks.
\newblock In {\em European Conference on Computer Vision}, 2020.

\bibitem{Restormer}
Syed~Waqas Zamir, Aditya Arora, Salman Khan, Munawar Hayat, Fahad~Shahbaz Khan,
  and Ming-Hsuan Yang.
\newblock Restormer: Efficient transformer for high-resolution image
  restoration.
\newblock In {\em IEEE Conference on Computer Vision and Pattern Recognition},
  2022.

\bibitem{MPRNet}
Syed~Waqas Zamir, Aditya Arora, Salman Khan, Munawar Hayat, Fahad~Shahbaz Khan,
  Ming-Hsuan Yang, and Ling Shao.
\newblock Multi-stage progressive image restoration.
\newblock In {\em IEEE Conference on Computer Vision and Pattern Recognition},
  2021.

\bibitem{nighttime_fast}
Jing Zhang, Yang Cao, Shuai Feng, Yu~Kang, and Chang~Wen Chen.
\newblock Fast haze removal for nighttime image using maximum reflectance
  prior.
\newblock In {\em IEEE Conference on Computer Vision and Pattern Recognition},
  2017.

\bibitem{nighttime_zhang}
Jing Zhang, Yang Cao, Zhengjun Zha, and Dacheng Tao.
\newblock Nighttime dehazing with a synthetic benchmark.
\newblock In {\em ACM International Conference on Multimedia}, 2020.

\bibitem{lpips}
Richard Zhang, Phillip Isola, Alexei~A. Efros, Eli Shechtman, and Oliver Wang.
\newblock The unreasonable effectiveness of deep features as a perceptual
  metric.
\newblock In {\em IEEE Conference on Computer Vision and Pattern Recognition},
  2018.

\bibitem{reflectionk_zhang}
Xuaner Zhang, Ren Ng, and Qifeng Chen.
\newblock Single image reflection separation with perceptual losses.
\newblock In {\em IEEE Conference on Computer Vision and Pattern Recognition},
  2018.

\bibitem{pspnet}
Hengshuang Zhao, Jianping Shi, Xiaojuan Qi, Xiaogang Wang, and Jiaya Jia.
\newblock Pyramid scene parsing network.
\newblock In {\em IEEE Conference on Computer Vision and Pattern Recognition},
  2017.

\bibitem{zhou2022lednet}
Shangchen Zhou, Chongyi Li, and Chen Change~Loy.
\newblock {LEDNet}: Joint low-light enhancement and deblurring in the dark.
\newblock In {\em European Conference on Computer Vision}, 2022.

\bibitem{CycleGAN}
Junyan Zhu, Taesung Park, Phillip Isola, and Alexei~A. Efros.
\newblock Unpaired image-to-image translation using cycle-consistent
  adversarial networks.
\newblock In {\em IEEE International Conference on Computer Vision}, 2017.

\end{thebibliography}
}
%




\end{document}